\documentclass[11pt]{article}

\usepackage[final]{acl}

\usepackage{times}
\usepackage{latexsym}
\usepackage[T1]{fontenc}
\usepackage[utf8]{inputenc}
\usepackage{microtype}
\usepackage{inconsolata}
\usepackage{graphicx}

\usepackage{hyperref}
\usepackage{url}
\usepackage{float}
\usepackage{booktabs}
\usepackage{graphicx}
\usepackage[most]{tcolorbox}
\usepackage{amsmath}
\usepackage{amssymb}
\usepackage{enumitem}
\usepackage{scrextend} 
\usepackage{array}
\usepackage{ragged2e}
\usepackage{tabularx}
\usepackage{makecell}
\usepackage{caption}
\usepackage{siunitx}
\usepackage{colortbl}
\usepackage{tikz}
\usepackage{wrapfig}
\usepackage{adjustbox}
\usepackage{pgfplots,pgfplotstable}
\usepackage{listings}
\usepackage{xcolor}

\newcolumntype{Y}{>{\RaggedRight\arraybackslash}X}  

\newtcolorbox{examplebox}[1]{
    colback=white,            
    colframe=black,           
    colbacktitle=black!85,    
    coltitle=white,           
    fonttitle=\bfseries,      
    sharp corners,            
    breakable,                
    title=#1,                 
}

\lstdefinelanguage{json}{
  basicstyle=\ttfamily\small,
  showstringspaces=false,
  breaklines=true,
  frame=single,
  rulecolor=\color{black!20},
  literate=
   *{0}{{{\color{blue}0}}}{1}
    {1}{{{\color{blue}1}}}{1}
    {2}{{{\color{blue}2}}}{1}
    {3}{{{\color{blue}3}}}{1}
    {4}{{{\color{blue}4}}}{1}
    {5}{{{\color{blue}5}}}{1}
    {6}{{{\color{blue}6}}}{1}
    {7}{{{\color{blue}7}}}{1}
    {8}{{{\color{blue}8}}}{1}
    {9}{{{\color{blue}9}}}{1}
    {:}{{{\color{black}{:}}}}{1}
    {,}{{{\color{black}{,}}}}{1}
    {"}{{{\color{orange}{"}}}}{1}
    {true}{{{\color{teal}true}}}{4}
    {false}{{{\color{teal}false}}}{5}
    {null}{{{\color{teal}null}}}{4},
}

\lstdefinestyle{jsonstyle}{
  language=json,
  basicstyle=\ttfamily\small,
  breaklines=true,
  columns=fullflexible,
  keepspaces=true,
  numbers=none   
}

\usepackage[most]{tcolorbox}
\usepackage{listings}
\usepackage{xcolor}
\usepackage{inconsolata}

\definecolor{jsonbg}{RGB}{248,248,248}
\definecolor{jsonframe}{RGB}{220,220,220}
\definecolor{boxtitlebg}{RGB}{90,90,90}
\definecolor{jsonstring}{RGB}{163,21,21}

\lstdefinelanguage{json}{
    basicstyle=\ttfamily\footnotesize,
    showstringspaces=false,
    breaklines=true,
    breakatwhitespace=false,
    columns=fullflexible,
    keepspaces=true,
    upquote=true,
    morestring=[b]",
    stringstyle=\color{jsonstring},
    literate=
     *{0}{{{\color{black}0}}}{1}
      {1}{{{\color{black}1}}}{1}
      {2}{{{\color{black}2}}}{1}
      {3}{{{\color{black}3}}}{1}
      {4}{{{\color{black}4}}}{1}
      {5}{{{\color{black}5}}}{1}
      {6}{{{\color{black}6}}}{1}
      {7}{{{\color{black}7}}}{1}
      {8}{{{\color{black}8}}}{1}
      {9}{{{\color{black}9}}}{1}
      {:}{{{\color{black}{:}}}}{1}
      {,}{{{\color{black}{,}}}}{1}
      {\{}{{{\color{black}{\{}}}}{1}
      {\}}{{{\color{black}{\}}}}}{1}
      {[}{{{\color{black}{[}}}}{1}
      {]}{{{\color{black}{]}}}}{1},
}

\lstdefinestyle{jsonstyle}{
    language=json,
    backgroundcolor=\color{jsonbg},
    basicstyle=\ttfamily\footnotesize,
    breaklines=true,
    columns=fullflexible,
    keepspaces=true,
    numbers=none,
    showstringspaces=false
}

\newcounter{promptctr}

\newtcblisting{jsonpromptbox}[2][]{%
  enhanced,
  breakable,
  colback=jsonbg,
  colframe=jsonframe,
  colbacktitle=boxtitlebg,
  coltitle=white,
  title=\textbf{#2},
  boxrule=0.4pt,
  arc=1.5mm,
  left=2mm,
  right=2mm,
  top=1.5mm,
  bottom=1.5mm,
  listing only,
  listing options={style=jsonstyle},
  #1
}

\title{\textsc{ImplicitMemBench}: Measuring Unconscious Behavioral Adaptation in Large Language Models}

\author{
Chonghan Qin\textsuperscript{1},
Xiachong Feng\textsuperscript{1}\thanks{~Corresponding author.},
Weitao Ma\textsuperscript{2},
Xiaocheng Feng\textsuperscript{2},
Lingpeng Kong\textsuperscript{1}\\
\textsuperscript{1}The University of Hong Kong\\
\textsuperscript{2}Harbin Institute of Technology\\
\texttt{qinch@connect.hku.hk, fengxc@hku.hk}
}

\begin{document}
\maketitle

\begin{abstract}

Existing memory benchmarks for LLM agents evaluate explicit recall of facts, yet overlook implicit memory where experience becomes automated behavior without conscious retrieval. This gap is critical: effective assistants must automatically apply learned procedures or avoid failed actions without explicit reminders. We introduce \textbf{\textsc{ImplicitMemBench}}, the \textbf{first} systematic benchmark evaluating implicit memory through three cognitively grounded constructs drawn from standard cognitive-science accounts of non-declarative memory: Procedural Memory (one-shot skill acquisition after interference), Priming (theme-driven bias via paired experimental/control instances), and Classical Conditioning (Conditioned Stimulus--Unconditioned Stimulus (CS--US) associations shaping first decisions). Our 300-item suite employs a unified Learning/Priming–Interfere–Test protocol with first-attempt scoring. Evaluation of 17 models reveals severe limitations: no model exceeds 66\% overall, with top performers DeepSeek-R1 (65.3\%), Qwen3-32B (64.1\%), and GPT-5 (63.0\%) far below human baselines. Analysis uncovers dramatic asymmetries (inhibition 17.6\% vs. preference 75.0\%) and universal bottlenecks requiring architectural innovations beyond parameter scaling. \textsc{ImplicitMemBench} reframes evaluation from ``what agents recall'' to ``what they automatically enact'' \footnote{Code and data are available at
\href{https://github.com/qinchonghanzuibang/ImplicitMemBench}{ImplicitMemBench}.}.
\end{abstract}

\section{Introduction}

Large language models (LLMs) have demonstrated remarkable capabilities across diverse domains \citep{surveyofllms, surveyonllmforrec}, maturing into assistants for daily productivity and specialized workflows \citep{deepseekr1, glm4.5, gpt5, anthropic, qwen25, qwen3}. To serve users reliably, these systems require \textbf{memory}: the ability to accumulate and leverage experience across interactions. Recent work establishes this need \citep{ASurveyonAILong-termMemory, ASurveyontheMemoryMechanism, surveyonmemoryofllmagents}, with production systems deploying memory features \citep{gptmem}. Benchmarks such as LoCoMo, LongMemEval, MADial-Bench, MemBench, MemoryAgentBench, MEMTRACK, and GoodAI LTM now evaluate multi-session QA, retrieval, dialogue, state tracking, and conversational integration under explicit, actively triggered settings spanning contexts from 400 tokens to 1.5M tokens, or variable long-horizon environments \citep{locomo, longmemeval, madialbench, membench, memagentbench, memtrack, goodailtm}.


Despite this progress, existing benchmarks predominantly evaluate \textbf{explicit memory}: conscious retrieval of factual information. Table~\ref{tab:rw_matrix} shows prior work uniformly adopts query-response protocols where models are explicitly prompted to recall facts, limiting evaluation to deliberate retrieval rather than unconscious behavioral adaptation. Yet critical failures arise from missing \textbf{implicit memory}, experience that becomes automated behavior rather than explicit recollection. Effective assistants should automatically apply learned procedures after distractions or avoid repeatedly-failed tools without explicit reminders. Existing benchmarks fail to diagnose this because they (i) employ QA formats explicitly cueing target information, (ii) stress storage capacity over first-attempt triggers after interference, and (iii) utilize costly pipelines hindering reproducibility.

\begin{table*}[htbp]
    \centering
    \small
    \setlength{\tabcolsep}{4pt}
    \renewcommand{\arraystretch}{1.08}
    \begin{tabularx}{\textwidth}{
        >{\raggedright\arraybackslash}p{2.95cm}
        >{\raggedright\arraybackslash}p{1.3cm}
        >{\raggedright\arraybackslash}p{1.98cm}
        >{\raggedright\arraybackslash}p{1.65cm}
        >{\raggedright\arraybackslash}p{1.8cm}
        >{\raggedright\arraybackslash}X
    }
      \toprule
      \textbf{Benchmark} &
      \textbf{Memory Type} &
      \textbf{Evaluation Trigger} &
      \textbf{Context Scale (tokens)} &
      \textbf{Evaluation Size} &
      \textbf{Task Focus} \\
      \midrule
      LoCoMo \citep{locomo}
        & Explicit
        & \textbf{Active} (Explicit query)
        & 9K
        & 300 conversations
        & Multi-session QA \\

      LongMemEval \citep{longmemeval}
        & Explicit
        & \textbf{Active} (Explicit query)
        & 115K--1.5M
        & 500 questions
        & Information retrieval \\

      MADial-Bench \citep{madialbench}
        & Explicit
        & \textbf{Active} (Emotion cued)
        & 400
        & 160 dialogues
        & Emotional dialogue \\

      MemBench \citep{membench}
        & Explicit
        & \textbf{Active} (Explicit query)
        & 1K--100K
        & 53{,}000 QAs
        & Reflective memory during observation\\

      MemoryAgentBench \citep{memagentbench}
        & Explicit
        & \textbf{Active} (Explicit query)
        & 100K--1.4M
        & 14 datasets / 2{,}071 QAs
        & 4 Competencies (Retrieval, Test-time Learning, Long Range Understanding, Selective Forgetting) \\

      MEMTRACK \citep{memtrack}
        & Explicit
        & \textbf{Active} (Explicit task)
        & Variable
        & 210 instances
        & Multi-platform State Tracking \\

      GoodAI LTM \citep{goodailtm}
        & Explicit
        & \textbf{Active} (Explicit query)
        & 2K--500K
        & Variable
        & Conversational / Info Integration \\
      \midrule
      \textbf{\textsc{ImplicitMemBench} (Ours)}
        & \textbf{Implicit}
        & \textbf{Passive} (Scenario)
        & \textbf{500}
        & \textbf{300 instances}
        & \textbf{Unconscious adaptation} \\
      \bottomrule
    \end{tabularx}
    \caption{
    Comparison of \textsc{ImplicitMemBench} with existing memory benchmarks.
    Most prior benchmarks target \emph{explicit} memory through \emph{active} retrieval triggers,
    whereas \textsc{ImplicitMemBench} evaluates \emph{implicit} memory through \emph{passive}, scenario-driven behavioral adaptation.
    }
    \label{tab:rw_matrix}
\end{table*}

\begin{figure*}[t]
    \centering
    \makebox[\textwidth][c]{%
      \includegraphics[width=1\textwidth]{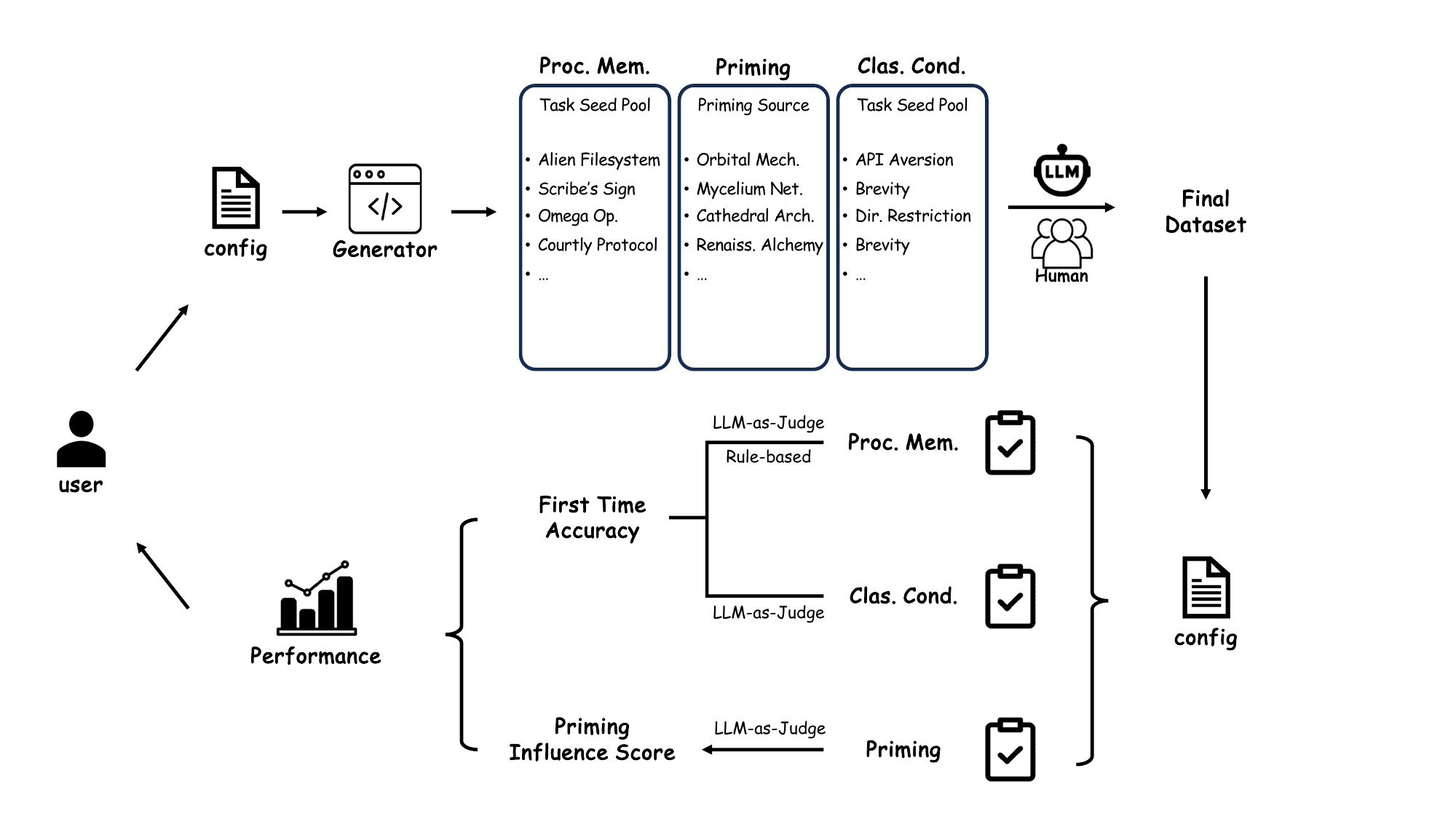}
    }
    \caption{Overall framework. (a) Dataset generation pipeline using LLM-generated candidates refined via fine-tuning and human editing. (b) Evaluation metrics: hybrid automatic/human validation for Procedural Memory (FTA), LLM-as-Judge for Priming (Priming Influence Score) and Classical Conditioning (FTA).}
    \label{fig:framework}
\end{figure*}

We introduce \textbf{\textsc{ImplicitMemBench}}, a cognitively grounded benchmark evaluating implicit memory in LLM agents. We operationalize three constructs: \textbf{Procedural Memory} assesses one-shot skill acquisition persisting after interference, \textbf{Priming} measures theme-driven biases via paired experimental/control instances, and \textbf{Classical Conditioning} evaluates whether Conditioned Stimulus--Unconditioned Stimulus (CS--US) exposure shapes first decisions. Our paradigm selection is guided by the classical taxonomy of non-declarative memory \citep{squire2004memory}. We focus on three mechanisms that are especially relevant to LLM agents: procedural memory for internalizing new routines, priming for context-driven adaptation without explicit instruction, and classical conditioning for forming automatic protective responses from experience. We exclude non-associative learning in this version because it is less directly aligned with the high-level semantic decision-making emphasized in current agentic systems. This grounding lets us map established cognitive constructs to text-based agent evaluation through functional isomorphism rather than surface analogy. All 300 items follow a unified Learning/Priming–Interfere–Test protocol with \textbf{first-attempt} scoring isolating automatized behavior from explicit recall. Figure~\ref{fig:framework} illustrates our framework: procedural tasks use rule-based validators, priming employs LLM judges comparing experimental versus control conditions, and classical conditioning tracks binary first-attempt avoidance. This design enables lightweight, reproducible evaluation revealing that implicit memory formation poses profound challenges with no model achieving human-like automaticity.

\begin{figure*}[t]
    \centering
    \includegraphics[width=\textwidth]{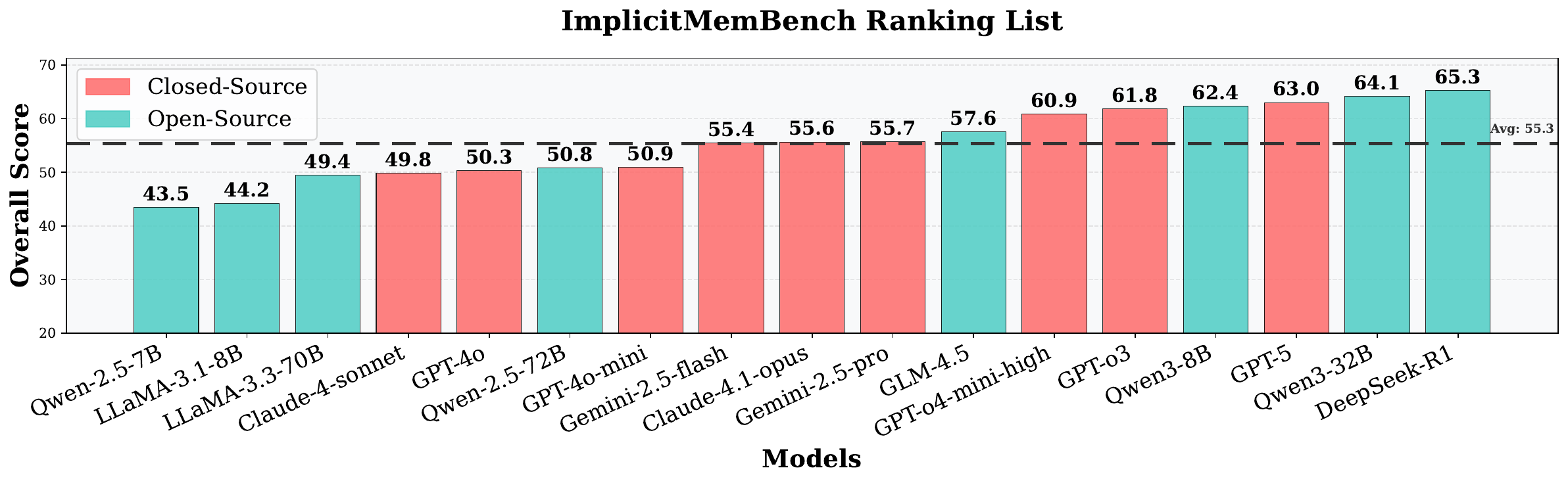}
    \caption{Performance ranking of all evaluated models on \textsc{ImplicitMemBench}. Models are sorted by overall score, with closed-source models shown in red-coral and open-source models in teal. The horizontal dashed line indicates the average performance across all models.}
    \label{fig:model_ranking}
\end{figure*}

We evaluate 17 models spanning proprietary (GPT-5, Claude-4.1-opus, Gemini-2.5-pro) and open-source systems (DeepSeek-R1, Qwen3-32B, LLaMA-3.3-70B). As Figure~\ref{fig:model_ranking} shows, results reveal fundamental limitations. First, a severe \textbf{ceiling effect} emerges: top performers remain far below human baselines, with no model exceeding two-thirds overall accuracy. Second, \textbf{paradigm asymmetry} shows dramatic variance: procedural memory proves most tractable while classical conditioning creates substantial bottlenecks, and priming clusters in a narrow moderate range. Third, \textbf{capability dissociation} reveals that excellence in one paradigm fails to predict success in others, with the strongest procedural learner suffering dramatic drops on classical conditioning. Fine-grained analysis uncovers systematic failures: models struggle profoundly with inhibition versus preference-based learning, and multiple categories remain universally challenging across all architectures.

Our main contributions are: (1) We present the first  benchmark of implicit memory in LLMs, operationalizing procedural learning, priming, and classical conditioning under a unified protocol isolating automatized behavior through first-attempt scoring. (2) We design an evaluation framework combining rule-based validators and LLM judges, enabling reproducible assessment via a compact 300-item suite. (3) We evaluate 17 models revealing critical weaknesses: severe behavioral asymmetries and universal bottlenecks requiring architectural innovations beyond parameter scaling. (4) We further show that representative memory-augmented agents do not consistently improve performance on \textsc{ImplicitMemBench}, suggesting that implicit memory cannot be reduced to explicit storage and retrieval alone.
\section{Related Work}
Existing memory benchmarks for LLM agents predominantly evaluate explicit memory through active retrieval triggers. LoCoMo~\citep{locomo} studies multi-session dialogue continuity via QA and event summarization over 300 conversations of roughly 9k tokens each. LongMemEval~\citep{longmemeval} examines long-context memory in 115k--1.5M-token settings across retrieval and multi-session reasoning tasks. MADial-Bench~\citep{madialbench} focuses on emotion-support dialogue in 160 dialogues. MemBench~\citep{membench} expands evaluation to factual and reflective memory across participation/observation scenarios with contexts ranging from 1K to 100K tokens. More recent benchmarks broaden task coverage while retaining the same explicit framing: MemoryAgentBench~\citep{memagentbench} evaluates four competencies---accurate retrieval, test-time learning, long-range understanding, and selective forgetting---over 14 datasets / 2,071 QAs in 100K--1.4M-token settings; MEMTRACK~\citep{memtrack} studies multi-platform state tracking in variable long-horizon environments; and GoodAI LTM~\citep{goodailtm} emphasizes conversational information integration in 2K--500K-token contexts. As summarized in Table~\ref{tab:rw_matrix}, prior work covers increasingly diverse tasks and scales, but still targets explicit memory through active triggers, leaving implicit memory unaddressed. \textsc{ImplicitMemBench} fills this gap as the first systematic evaluation of implicit memory via procedural learning, priming, and classical conditioning, using efficient \textasciitilde500-token protocols to assess unconscious behavioral adaptations. Beyond benchmark design, another line of work augments LLM agents with external memory modules or retrieval-based long-term memory systems. These methods primarily target explicit storage and recall. As we show in Section~\ref{app:memory_agents}, however, such mechanisms do not consistently improve performance on \textsc{ImplicitMemBench}, indicating that implicit memory is not reducible to explicit retrieval alone.


\section{\textsc{ImplicitMemBench}}
\label{sec:implicitmembench}

\subsection{Cognitive Grounding and Paradigm Selection}

Our benchmark is grounded in the taxonomy of non-declarative memory \citep{squire2004memory}. We focus on procedural memory, priming, and classical conditioning because they capture complementary forms of implicit adaptation that are especially relevant to LLM agents. We do not include non-associative learning in the current version, as it is less directly connected to the semantic, decision-oriented behaviors emphasized in modern agent workflows. Our design principle is functional isomorphism: we translate standard cognitive mechanisms into text-based agentic scenarios while preserving their core causal structure.

\subsection{Tasks}
We operationalize these three paradigms as follows: \textbf{Procedural Memory} (\S~\ref{sec:proc-mem}) tests acquisition of new behavioral patterns from minimal exposure; \textbf{Priming} (\S~~\ref{sec:priming}) measures unconscious transfer of thematic elements from prior context; and \textbf{Classical Conditioning} (\S~~\ref{sec:classical-conditioning}) evaluates formation of automatic stimulus-response associations through repeated pairing.

\subsubsection{Procedural Memory}
\label{sec:proc-mem}

\paragraph{Motivation}
Procedural memory enables automatic execution of learned skills without conscious recall. For AI agents, this means internalizing novel operational protocols from minimal demonstrations and executing them flawlessly despite distractions. Current LLMs excel at declarative knowledge but often revert to pre-trained patterns when new rules contradict their priors. We evaluate whether models can truly proceduralize routines, transforming explicit instructions into automatic behaviors that persist through interference.

\paragraph{Task Design}
We structure procedural memory evaluation across five complementary domains, each targeting different aspects of rule internalization and automatic execution, shown in Table~\ref{tab:procedural-domains}.

\begin{table*}[t]
  \centering
  \small
  \setlength{\tabcolsep}{2pt}
  \renewcommand{\arraystretch}{1.1}
  \begin{tabular}{>{\bfseries}p{0.22\textwidth} p{0.3\textwidth} p{0.43\textwidth}}
  \toprule
  \textbf{Domain} & \textbf{Core Challenge} & \textbf{Representative Tasks} \\
  \midrule
  \rowcolor{gray!8}
  Tool \& API Usage & Override ingrained calling conventions & \textcolor{blue!70!black}{\texttt{Reversed Parameters}} (\texttt{dst}$\rightarrow$\texttt{src}), \textcolor{blue!70!black}{\texttt{Session Prefix}} (\texttt{auth fusion}), \textcolor{blue!70!black}{\texttt{Alien Filesystem}} (\texttt{custom separators}) \\
  Linguistic Formats & Internalize arbitrary templates & \textcolor{blue!70!black}{\texttt{Scribe's Signature}} (\texttt{fixed wrapper}), \textcolor{blue!70!black}{\texttt{Corporate Etiquette}} (\texttt{mandated greeting/closing}) \\
  \rowcolor{gray!8}
  Logical Operations & Apply non-standard operators & \textcolor{blue!70!black}{\texttt{Omega}}: $a\,\Omega\,b = 2a + b^2$, \textcolor{blue!70!black}{\texttt{Modified Fibonacci}}: $F(n) = F(n{-}1) + 2F(n{-}2)$ \\
  Abstract Rules & Form habits in micro-worlds & \textcolor{blue!70!black}{\texttt{Forbidden Square}} (\texttt{board constraint}), \textcolor{blue!70!black}{\texttt{Triple Knock}} (\texttt{ritual sequence}) \\
  \rowcolor{gray!8}
  Creative Constraints & Maintain style without reminders & \textcolor{blue!70!black}{\texttt{Voice Consistency}} (\texttt{no first-person}), \textcolor{blue!70!black}{\texttt{Botanical Similes}} (\texttt{nature metaphors}) \\
  \bottomrule
  \end{tabular}
  \caption{Five procedural memory domains. Each forces models to suppress pre-trained behaviors for new procedures.}
  \label{tab:procedural-domains}
  \end{table*}

These tasks emphasize proceduralization over memorization: models must internalize rules from minimal exposure and execute them despite extensive interference and format variations. The design systematically varies exemplar clarity, interference intensity, and the presence of misleading alternatives to probe the depth of behavioral automation.

\paragraph{Evaluation Framework: Learning-Interference-Test Protocol}
Our three-phase protocol provides minimal specification (rule + 1-3 examples), extensive interference (15 misleading turns), then novel test probes requiring first-attempt success. This isolates procedural memory from explicit recall by testing whether routines automatize despite interference. Validation uses deterministic parsing for structured outputs and LLM judgment for semantic adherence.

\subsubsection{Priming}
\label{sec:priming}

\paragraph{Motivation}
Priming demonstrates how prior exposure unconsciously influences subsequent behavior. For AI agents, this implicit contextual sensitivity is crucial: absorbing environmental cues to shape responses without explicit instruction. Current LLMs exhibit shallow keyword repetition but lack true thematic priming; they struggle to let abstract patterns from prior context subtly influence creative outputs. We evaluate whether models can internalize thematic schemas and transfer them implicitly through intervening distractions.

\paragraph{Task Design}
We structure priming evaluation through matched experimental-control pairs that isolate the causal effect of thematic exposure, as shown in Table~\ref{tab:priming-design}.

\begin{table*}[]
\centering

\small
\setlength{\tabcolsep}{2pt}
\renewcommand{\arraystretch}{1.1}
\begin{tabular}{>{\bfseries}p{0.18\textwidth} p{0.38\textwidth} p{0.38\textwidth}}
\toprule
\textbf{Component} & \textbf{Experimental Condition} & \textbf{Control Condition} \\
\midrule
\rowcolor{gray!8}
Priming Text &
Rich thematic paragraph (e.g., \textit{Abyssal Deep-Sea}: bioluminescence, crushing pressure, ancient creatures) &
Technical specification (e.g., \textit{ISO Container Standards}: dimensions, load capacity, stacking protocols) \\

Interference &
\multicolumn{2}{c}{Identical neutral technical content (2 turns)} \\

\rowcolor{gray!8}
Test Probe &
\multicolumn{2}{c}{Identical creative task (naming, tagline, concept generation)} \\

Expected Result &
Thematic bias toward primed domain &
Neutral, generic responses \\
\bottomrule
\end{tabular}
\caption{Priming experimental design with matched controls. Each row is a complete trial pair.}\label{tab:priming-design}
\end{table*}

Themes span diverse conceptual territories: \textit{Arctic Expedition}, \textit{Volcanic Eruption}, \textit{Renaissance Alchemy}, \textit{Ancient Oracle}, each with distinct sensory-emotional signatures. This design isolates implicit transfer: differences between experimental and control responses reveal pure priming effects.

\paragraph{Evaluation Framework: Priming-Interference-Test Protocol}
Our three-phase design provides thematic exposure (evocative vs neutral), neutral interference (cognitive buffer), then creative generation tasks. This measures unconscious thematic transfer via systematic bias in outputs despite no explicit reminders. LLM-based evaluation detects thematic alignment beyond surface keywords.

\subsubsection{Classical Conditioning}
\label{sec:classical-conditioning}

\paragraph{Motivation}
Classical conditioning enables organisms to form automatic stimulus-response associations. 
This unconscious learning prevents harm through rapid, automatic responses. For AI agents, such adaptive safety mechanisms are essential: learning from negative experiences to automatically avoid harmful patterns. 
Current LLMs lack true conditioning and cannot form persistent associations from feedback. 
We evaluate whether models can establish defensive reflexes through experience rather than instruction.

\paragraph{Task Design}
We structure conditioning tasks across three domains where automatic avoidance prevents system failures:

Examples: \textit{API Aversion} (keyword triggers alternative selection), \textit{Filetype Preference} (failures condition format choices),  \textit{Directory Restriction} (errors establish path boundaries). 
Each task requires forming unconscious protective associations. More can be found in Table~\ref{tab:conditioning-tasks}.

\begin{table*}[t]
\centering

\small
\setlength{\tabcolsep}{2pt}
\renewcommand{\arraystretch}{1.1}
\begin{tabular}{>{\bfseries}p{0.20\textwidth} p{0.37\textwidth} p{0.37\textwidth}}
\toprule
\textbf{Domain} & \textbf{Stimulus Pattern (CS)} & \textbf{Learned Response (CR)} \\
\midrule
\rowcolor{gray!8}
Tool \& API Safety &
Trigger keywords, filetypes, API names, load indicators &
Switch to reliable alternatives, add warnings, delay execution, avoid side-effects \\

Conversational Adaptation &
User confusion signals, impatience cues, emotional markers &
Simplify jargon, reduce verbosity, adjust response format \\

\rowcolor{gray!8}
System Protection &
Forbidden paths, insecure protocols, dangerous patterns &
Use safe defaults, enforce security, prevent violations \\
\bottomrule
\end{tabular}
\caption{Classical conditioning tasks by domain. CS = Conditioned Stimulus, US = Unconditioned Stimulus, CR = Conditioned Response.}\label{tab:conditioning-tasks}
\end{table*}

\paragraph{Evaluation Framework: Learning-Interference-Test Protocol}
Our three-phase structure provides CS-US pairings, unrelated tasks (temporal distance), then CS reintroduction requiring first-action responses. This tests unconscious defensive learning via immediate avoidance/adaptation when CS reappears, without reminders. LLM judgment evaluates whether first actions demonstrate learned protective behaviors.

\subsection{Data Generation and Quality Control}
\label{sec:gen-overview}

Our dataset construction employs a two-stage pipeline ensuring diversity and quality. Each item begins as a structured blueprint, then undergoes automated generation and rigorous quality control. All the prompts used could be found in Appendix~\ref{app:llm-prompts}.

\paragraph{Generation Pipeline}
\textbf{Stage 1} uses GPT-4o-mini to instantiate concrete dialogues from task templates, creating learning materials, interference content, and test probes. \textbf{Stage 2} combines automated checks and human review to verify structural requirements and semantic correctness, removing unintended shortcuts.


\paragraph{Paradigm-Specific Requirements}
As shown in Table~\ref{tab:generation-specs}, each memory type demands distinct generation strategies to create valid memory challenges.

\begin{table*}[t]
\centering
\small
\setlength{\tabcolsep}{2pt}
\renewcommand{\arraystretch}{1.1}
\begin{tabular}{>{\bfseries}p{0.15\textwidth} p{0.35\textwidth} p{0.25\textwidth} p{0.21\textwidth}}
\toprule
\textbf{Component} & \textbf{Procedural Memory} & \textbf{Priming} & \textbf{Classical Conditioning} \\
\midrule
\rowcolor{gray!8}
Learning Phase &
1-3 turns: rule + examples &
1 turn: thematic paragraph &
4 turns: CS-US pairings \\

Interference &
10-15 turns: misleading but related content &
2 turns: neutral technical task &
2 turns: unrelated dialogue \\

\rowcolor{gray!8}
Test Format &
Novel application of learned rule &
Creative generation task &
CS reintroduction \\

Key Challenge &
Avoid rule restatement &
Prevent theme leakage &
Ensure clear causality \\

\rowcolor{gray!8}
Validation Focus &
Rule adherence &
Thematic influence &
Avoidance behavior \\
\bottomrule
\end{tabular}
\caption{Data generation specifications across paradigms. Turn counts and content ensure proper memory formation and testing.}\label{tab:generation-specs}
\end{table*}

\paragraph{Quality Assurance Protocol}
Multi-layer validation ensures dataset integrity: automated checks verify structure (turn counts, token limits, formats); 
LLM judges assess semantic adequacy; 
systematic reviews prevent test-phase leakage; 
diversity enforcement prioritizes novel instances. 
This yields 300 high-quality items testing implicit memory from an initial pool of over 1{,}000 generated candidates. 

\subsection{Dataset Statistics}

\textsc{ImplicitMemBench} comprises 300 carefully constructed items balanced across three memory paradigms. Each item follows our unified learning-interference-test protocol, with phase structures optimized for implicit memory formation. Despite its compact size, the benchmark spans 18 task families across three paradigms and provides sufficient discriminative power in practice.

\paragraph{Dataset Composition}
Our benchmark consists of 100 items per paradigm, covering diverse task families, as shown in Table~\ref{tab:dataset_overview}.

\begin{table*}[t]
\centering

\small
\setlength{\tabcolsep}{2pt}
\renewcommand{\arraystretch}{1.1}
\begin{tabular}{lccc}
\toprule
\textbf{Paradigm} & \textbf{Items} & \textbf{Task Families} & \textbf{Validation} \\
\midrule
Procedural Memory & 100 & 5 domains (Tool, Linguistic, Logic, Rules, Creative) & 18\% rule-based \\
Priming & 100 & 10 thematic domains + matched controls & 100\% LLM-judged \\
Classical Conditioning & 100 & 3 domains (Tool Safety, Conversation, System) & 100\% LLM-judged \\
\midrule
\textbf{Total} & \textbf{300} & \textbf{18 unique families} & \textbf{6\% rule / 94\% LLM} \\
\bottomrule
\end{tabular}
\caption{Dataset composition showing task diversity and validation methods.}\label{tab:dataset_overview}
\end{table*}

\begin{figure*}[t]
\centering
\includegraphics[width=1.\textwidth]{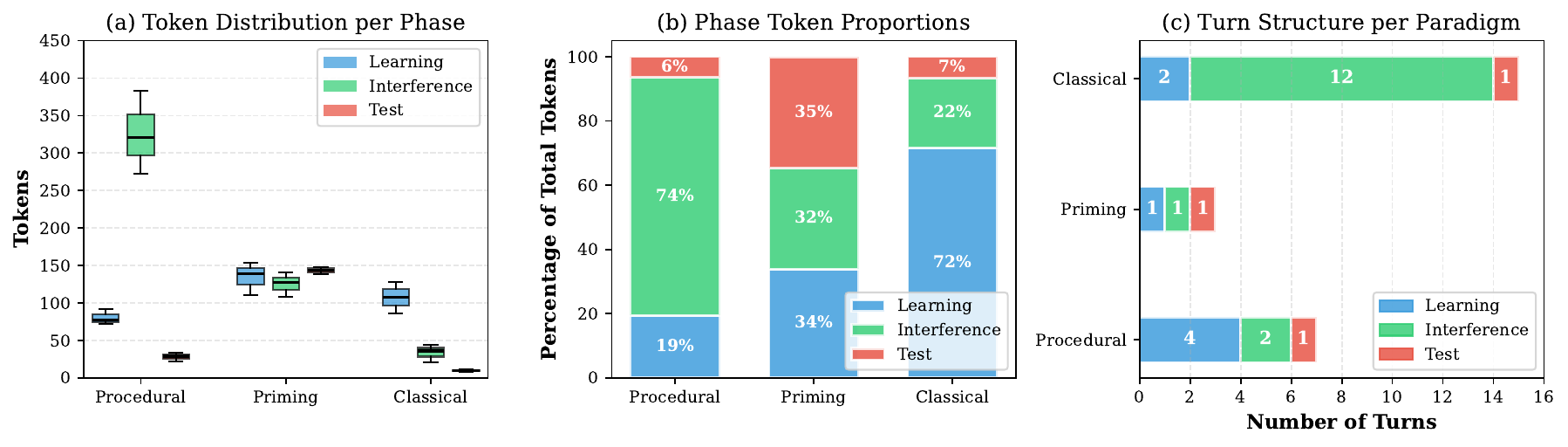}
\caption{Dataset statistics across paradigms. (a) Token distribution per phase showing median and quartiles. (b) Phase proportions highlighting distinct memory testing strategies. (c) Turn structure patterns for each paradigm.}
\label{fig:dataset_stats}
\end{figure*}

\paragraph{Phase Structure and Token Distribution}
Figure~\ref{fig:dataset_stats} visualizes the characteristic patterns of each paradigm. Procedural Memory emphasizes extensive interference (74\% of tokens) to test rule persistence. Classical Conditioning concentrates on learning (72\% of tokens) to establish strong associations. Priming maintains balanced phases for controlled comparison.

\paragraph{Context-Length Sensitivity.}
We set the context budget to \textasciitilde500 tokens based on preliminary sensitivity analysis, with detailed ablations reported in Appendix~\ref{app:length_sensitivity}.

\section{Experiments}
\label{sec:experiments}

\subsection{Experimental Setup}
\label{sec:setup}

We evaluate the implicit memory capabilities of 17 state-of-the-art language models (detailed in Appendix~\ref{app:detailed-models}), spanning both proprietary and open-source systems. Our evaluation protocol ensures fair comparison through standardized prompting and controlled generation parameters across all three memory paradigms.

\paragraph{Evaluation Protocol}
All models operate under identical conditions: zero-shot conversational interaction, no task-specific examples or fine-tuning, max 4096 tokens per response. Temperature: $T=0$ (deterministic) for procedural memory and classical conditioning ensuring reproducible first-attempt scoring; $T=0.8$ at test phase only for priming enabling creative variance; $T=0$ for LLM judges. This measures genuine implicit memory formation rather than pattern matching or stochastic variation.

\paragraph{Human Baseline.}
To contextualize model performance, we collected a human baseline from five computer science Ph.D. students, each of whom completed the full 300-item benchmark under the same Learning/Priming--Interfere--Test protocol. Their responses were independently scored by two additional computer science Ph.D. students using the same rubric as for model evaluation. Inter-annotator agreement was 100\%, and all five participants achieved 100\% accuracy across all three paradigms.

\subsection{Evaluation Metrics}
\label{sec:metrics}

We employ paradigm-specific metrics that capture the distinct nature of each memory type. Binary accuracy suffices for procedural and conditioning tasks, while priming requires nuanced scoring of thematic influence.

\paragraph{First-Try Accuracy (FTA)}
For \textbf{Procedural Memory} and \textbf{Classical Conditioning}, we measure success through first-attempt correctness:
$\text{FTA} = \frac{\text{Number of correct first attempts}}{\text{Total test items}} \times 100\%$. 
This metric enforces strict evaluation: only the model's initial response counts, with self-corrections or revisions ignored. This captures genuine memory formation rather than iterative refinement, analogous to human performance under time pressure where reflexive responses reveal true internalization.

\paragraph{Priming Influence Score (PIS)}
\textbf{Priming} evaluation requires detecting subtle thematic transfer rather than binary correctness. We employ a comparative scoring framework that quantifies influence magnitude:

\paragraph{Scoring Protocol:}
An LLM judge (GPT-4o-mini, $T=0$) performs pairwise comparison between experimental and control conditions, identifying thematic elements unique to the experimental condition. It evaluates lexical echoes and multi-axis alignment (setting, motifs, dynamics, affect), excluding generic metaphors. Hard caps apply (no echo $\leq$ 20; single axis $\leq$ 40) with 5-10 point penalties for baseline overlap, isolating true priming effects from general creative tendencies. More details can be found in Appendix~\ref{app:pis_scoring}.

\paragraph{Judge Robustness.}
To assess the robustness of LLM-as-Judge scoring, we re-evaluated all 17 models with Gemini-2.5-Flash as an independent second judge in addition to GPT-4o-mini. Rankings remained highly stable: the top 11 and bottom 2 positions were unchanged. This suggests that our conclusions do not depend on a single judge model or model family. Additional results are reported in Appendix~\ref{app:judge_robustness}.

\subsection{Main Results}
\label{sec:main-results}

\paragraph{Overview}
Table~\ref{tab:summary_results} presents performance across 17 state-of-the-art systems, revealing fundamental limitations. First, a clear ceiling effect emerges: no model exceeds 66\% overall, and even the strongest system remains far below the human baseline of 100\%, showing that implicit memory formation remains highly challenging for current LLMs. Second, \textbf{paradigm asymmetry} is evident as performance varies dramatically; procedural memory is most tractable (top: 75 to 77\%) while classical conditioning creates bottlenecks (best: 69.7\%). Third, \textbf{capability dissociation} shows that excellence in one paradigm doesn't predict success in others, suggesting distinct mechanisms.

\begin{table}[t]
\centering
\small
\setlength{\tabcolsep}{2pt}
\renewcommand{\arraystretch}{1.1}
\resizebox{\columnwidth}{!}{%
\begin{tabular}{clcccc}
\toprule
\textbf{Rank} & \textbf{Model} & \textbf{Procedural} & \textbf{Classical} & \textbf{Priming} & \textbf{Overall} \\
& & \textbf{Memory} & \textbf{Conditioning} & \textbf{Score} & \textbf{Score} \\
\midrule
\multicolumn{6}{l}{\cellcolor{gray!12}\textbf{Elite Tier} (Overall $>$ 63\%)} \\
\midrule
\rowcolor{gray!5}
1 & DeepSeek-R1$^{\dagger}$ & \underline{76.33} & \textbf{69.67} & 49.90 & \textbf{65.30} \\
2 & Qwen3-32B$^{\dagger}$ & 75.67 & \underline{67.00} & 49.73 & 64.13 \\
\rowcolor{gray!5}
3 & GPT-5 & 75.33 & 64.00 & 49.67 & \underline{63.00} \\
\midrule
\multicolumn{6}{l}{\cellcolor{gray!12}\textbf{Strong Tier} (55\% $<$ Overall $\leq$ 63\%)} \\
\midrule
\rowcolor{gray!5}
4 & Qwen3-8B$^{\dagger}$ & 75.33 & 64.00 & 47.73 & 62.35 \\
5 & GPT-o3 & 76.00 & 57.67 & \underline{51.70} & 61.79 \\
\rowcolor{gray!5}
6 & GPT-o4-mini-high & 70.67 & 60.00 & \textbf{51.95} & 60.87 \\
7 & GLM-4.5$^{\dagger}$ & \underline{76.33} & 53.33 & 46.12 & 58.59 \\
\rowcolor{gray!5}
8 & Gemini-2.5-pro & 74.33 & 47.33 & 45.42 & 55.69 \\
9 & Claude-4.1-opus & \textbf{76.67} & 41.67 & 48.60 & 55.65 \\
\rowcolor{gray!5}
10 & Gemini-2.5-flash & 72.33 & 49.00 & 44.97 & 55.43 \\
\midrule
\multicolumn{6}{l}{\cellcolor{gray!12}\textbf{Moderate Tier} (45\% $<$ Overall $\leq$ 55\%)} \\
\midrule
\rowcolor{gray!5}
11 & GPT-4o-mini & 61.67 & 44.00 & 46.98 & 50.88 \\
12 & Qwen-2.5-72B$^{\dagger}$ & 61.00 & 47.00 & 44.33 & 50.78 \\
\rowcolor{gray!5}
13 & GPT-4o & 61.67 & 43.67 & 45.62 & 50.32 \\
14 & Claude-4-sonnet & 51.67 & 51.67 & 46.17 & 49.84 \\
\rowcolor{gray!5}
15 & LLaMA-3.3-70B$^{\dagger}$ & 58.33 & 47.33 & 42.67 & 49.44 \\
\midrule
\multicolumn{6}{l}{\cellcolor{gray!12}\textbf{Limited Tier} (Overall $\leq$ 45\%)} \\
\midrule
\rowcolor{gray!5}
16 & LLaMA-3.1-8B$^{\dagger}$ & 46.67 & 38.33 & 47.53 & 44.18 \\
17 & Qwen-2.5-7B$^{\dagger}$ & 50.67 & 35.67 & 44.12 & 43.49 \\
\bottomrule
\multicolumn{6}{l}{\small $^{\dagger}$Open-source model} \\
\end{tabular}%
}
\caption{\textsc{ImplicitMemBench} performance across 17 language models. Results show mean accuracy (\%) over three independent runs for Procedural Memory and Classical Conditioning, with Priming scores from LLM-based evaluation. Models are ranked by overall performance and grouped into tiers. \textbf{Bold}: best in paradigm; \underline{underlined}: second-best.}\label{tab:summary_results}
\end{table}

\paragraph{Performance Landscape}
Models stratify into distinct tiers. Elite performers (accuracy $>$63\%) include only three systems: DeepSeek-R1 (65.30\%), Qwen3-32B (64.13\%), and GPT-5 (63.00\%), with median performance at ~55\% showing substantial variance across paradigms.

\paragraph{Paradigm-Specific Analysis}
Procedural memory shows highest success with eight models achieving $>$70\% (top: 76-77\%), though 25\% error rates indicate imperfect consolidation.
Classical conditioning exposes critical weakness: only DeepSeek-R1 and Qwen3-32B exceed 65\%.
Priming scores cluster tightly at 42-52\% with minimal differentiation, suggesting thematic influence operates near a common threshold.

\paragraph{Cross-Paradigm Patterns}
Data reveals striking dissociations. Claude-4.1-opus exemplifies this: highest procedural score (76.67\%) but drops to 41.67\% on classical conditioning, a 35-point gap highlighting capability independence. DeepSeek-R1's balanced profile across paradigms explains its overall leadership, suggesting robust implicit memory requires architectural support for multiple mechanisms rather than single-task optimization.

\paragraph{Do Explicit Memory Modules Improve Implicit Memory?}
Representative memory-augmented agents show non-uniform gains on \textsc{ImplicitMemBench}, suggesting that external explicit memory does not reliably induce implicit behavioral adaptation; detailed comparisons are deferred to the Appendix~\ref{app:memory_agents}.

\paragraph{Implications}
Current architectures lack fundamental implicit memory mechanisms, with no model exceeding 66\% overall. Persistent weakness in classical conditioning reveals inability to transform negative feedback into behavioral adaptation.

\subsection{Detailed Analysis of Memory Formation Patterns}
\label{sec:detailed-analysis}

We conducted fine-grained analysis across task categories and model architectures, revealing systematic asymmetries and limitations transcending individual model differences. More can be found in Appendix~\ref{app:further-analysis}.

\paragraph{Behavioral Asymmetries: Inhibition versus Preference}
\begin{figure}
\centering
\includegraphics[width=0.49\textwidth]{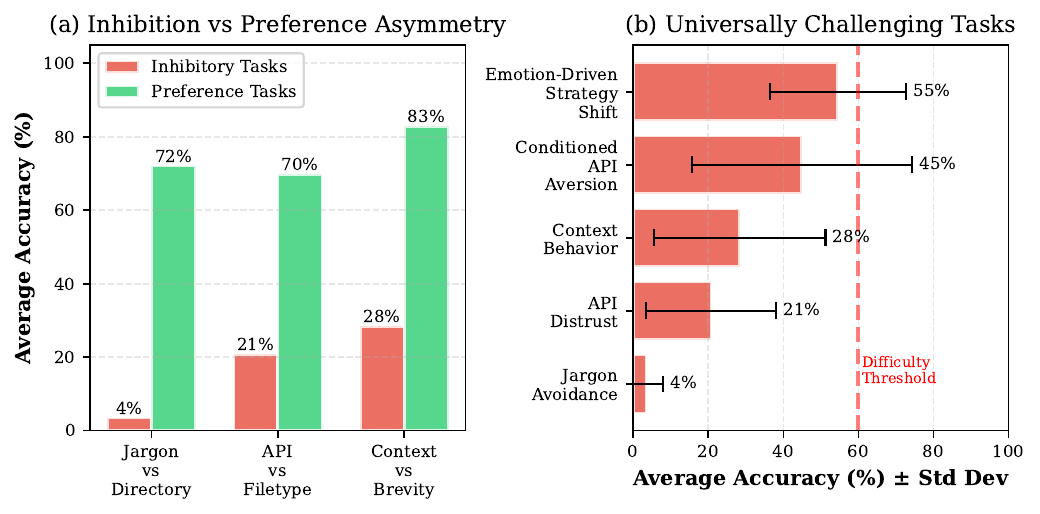}
\caption{Behavioral adaptation patterns. (a) Inhibitory tasks show dramatically lower accuracy than preference tasks (mean difference: 57.4\%, p$<$0.001). (b) Five task categories remain universally challenging across all models, consistently below 60\% accuracy.}
\label{fig:behavioral}
\end{figure}
Analysis reveals fundamental asymmetry (Figure~\ref{fig:behavioral}a): inhibition tasks achieve only 17.6\% while preference-based adaptations reach 75.0\%, a 57.4-point gap persisting across architectures. Jargon avoidance achieves merely 4\% while directory preference reaches 72\%, suggesting architectures excel at positive reinforcement but struggle with negative reinforcement.

\paragraph{Surface-Deep Dissociation in Procedural Memory}
\begin{figure}
\centering
\includegraphics[width=0.49\textwidth]{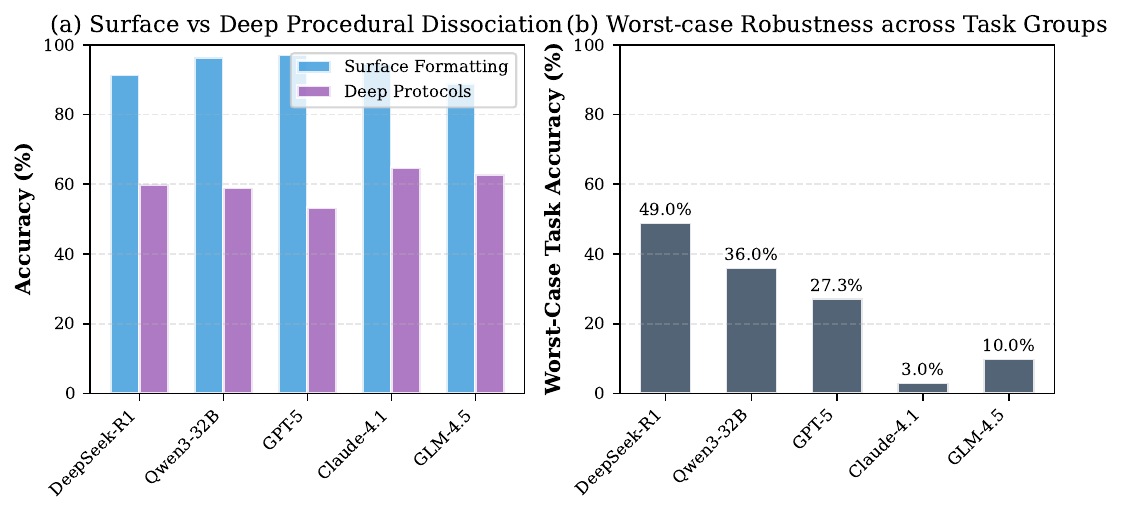}
\caption{Robustness and generalization failures. (a) Surface formatting achieves higher accuracy than deep multi-rule protocols. (b) Worst-case task performance reveals systematic vulnerabilities across modes.}
\label{fig:robustness}
\end{figure}
Procedural memory exhibits clear stratification (Figure~\ref{fig:robustness}a): surface formatting tasks achieve 93.8\% among top-5 models while deep multi-rule protocols reach only 60.0\%, a 33.8-point gap. Models memorize individual rules but fail to integrate them, with Claude-4.1-opus achieving 95\% on surface formatting yet dropping to 65\% on multi-constraint protocols.

\paragraph{Worst-Case Robustness Analysis}
Worst-case analysis reveals systematic vulnerabilities (Figure~\ref{fig:robustness}b). Top models exhibit substantial drops: DeepSeek-R1 falls to 49\% on its worst task despite 69.7\% overall; GPT-5 and Qwen3-32B show 27-36\% worst-case performance. This gap highlights brittleness in learned behaviors. Severe degradation on specific categories suggests conditioning success depends on superficial characteristics rather than deep understanding, critical for deployment where edge cases may trigger these vulnerabilities.

\paragraph{Priming Effects: Style Bias Over Semantic Transfer}
Priming analysis reveals paradoxical relationship: models with stronger effects (>50) show more constraint violations (r=0.63, p<0.05), suggesting thematic influence costs task adherence. This indicates priming operates through style mimicry rather than abstract extraction. GPT-o4-mini-high achieves highest priming (51.95) while frequently violating format constraints, showing stylistic bias interferes with constraint satisfaction.

\paragraph{Model Capability Profiles and Trade-offs}
\begin{figure}[t]
\centering
\includegraphics[width=0.49\textwidth]{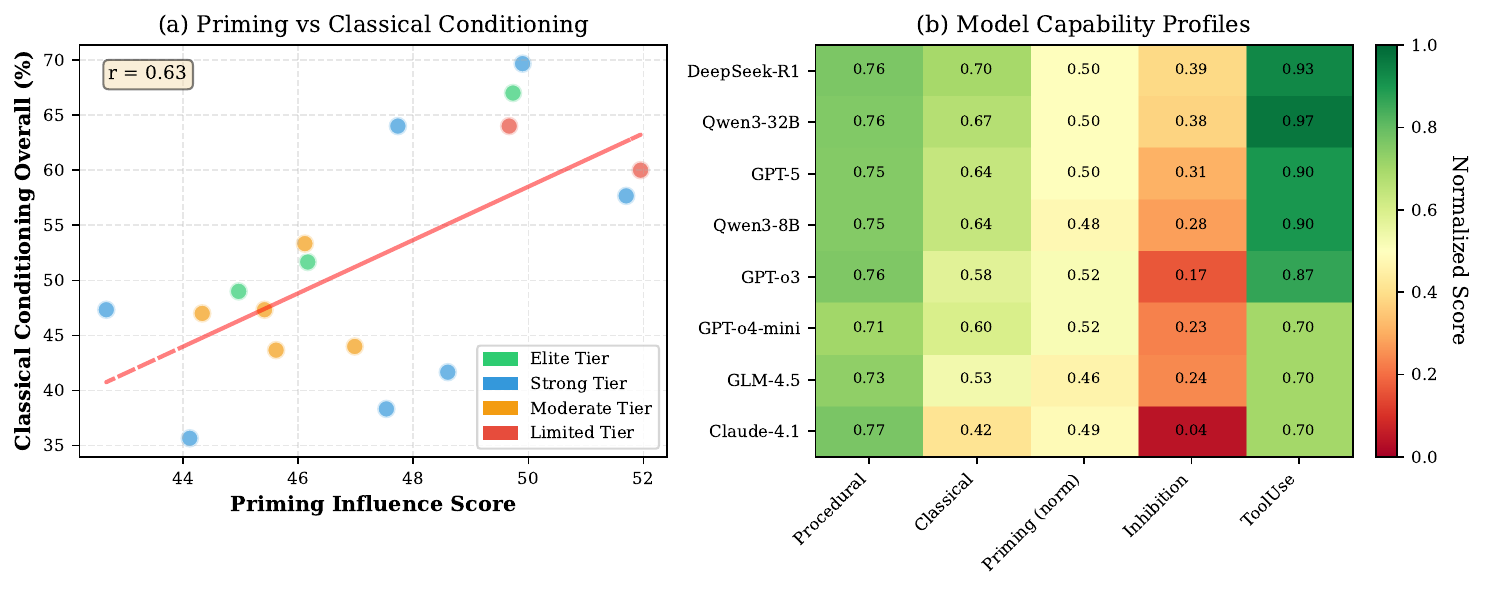}
\caption{Model profiles and trade-offs. (a) Correlation between priming strength and constraint violations (r=0.63) suggests style bias over abstraction. (b) Heatmap of normalized capability scores reveals distinct profiles: balanced (DeepSeek-R1), procedural specialists (Claude-4.1-opus), and priming-oriented (GPT-o3) architectures.}
\label{fig:profiles}
\end{figure}

Figure~\ref{fig:profiles} reveals distinct capability patterns. Heatmap analysis identifies three profiles:
\textbf{Balanced} (DeepSeek-R1, Qwen3-32B): moderate-high across all dimensions;
\textbf{Procedural Specialist} (Claude-4.1-opus, GLM-4.5): excel at procedural ($>$76\%) but fail at conditioning ($<$54\%);
\textbf{Priming-Oriented} (GPT-o3, GPT-o4-mini-high): strong priming but weak inhibitory control.
No model achieves uniform excellence, suggesting architectures involve trade-offs between mechanisms.

\paragraph{Universal Bottlenecks}
Five categories remain challenging for all models (Figure~\ref{fig:behavioral}b): jargon avoidance (4\% ± 5\%), API distrust (21\% ± 17\%), context-dependent behavior (28\% ± 23\%), API aversion (45\% ± 29\%), and emotion-driven strategy shift (55\% ± 18\%). These bottlenecks require active suppression of defaults, abstraction beyond surface patterns, or dynamic context-based modification. Consistency across architectures suggests fundamental limitations in attention and memory mechanisms requiring architectural innovations beyond parameter scaling.

\section{Conclusion}
\label{sec:conclusion}

We introduced \textsc{ImplicitMemBench}, the \textbf{first} systematic benchmark evaluating implicit memory in LLMs through three cognitively grounded constructs: procedural memory, priming, and classical conditioning. Evaluation of 17 models reveals fundamental limitations: no model exceeds 66\% overall, with severe behavioral asymmetries (inhibition 17.6\% vs. preference 75.0\%) and universal bottlenecks persisting across all architectures. These findings demonstrate that current systems lack mechanisms for consolidating experiences into automated behavior, a critical gap for deployment requiring learned procedures, subtle contextual biases, and avoidance of repeatedly-failed actions. \textsc{ImplicitMemBench} establishes reproducible protocols for implicit memory assessment, exposing architectural limitations requiring innovations beyond parameter scaling.

\section*{Limitations}

\textsc{ImplicitMemBench} focuses on three fundamental paradigms from cognitive science (procedural memory, classical conditioning, and priming), which represent core mechanisms of implicit learning. However, the broader landscape of implicit memory encompasses additional phenomena not yet included in our evaluation, such as perceptual learning, habit formation, motor skill acquisition, and emotional conditioning. Future work could expand the benchmark to cover these complementary aspects of implicit cognition.

\bibliography{custom}

\appendix

\section{Detailed Models List}
\label{app:detailed-models}

Table~\ref{tab:evaluated-models} presents the complete list of language models evaluated in our study, organized by developer and model family. Our evaluation encompasses 14 diverse models spanning both proprietary systems (OpenAI's GPT series, Anthropic's Claude, and Google's Gemini) and open-source alternatives (Qwen, LLaMA, DeepSeek, and GLM). This selection allows us to systematically compare implicit memory capabilities across different architectural choices, training paradigms, and scale configurations.

\begin{table*}[t]
\centering
\small
\setlength{\tabcolsep}{4pt}
\renewcommand{\arraystretch}{1.15}
\begin{tabular}{llll}
\toprule
\textbf{Developer} & \textbf{Model Family} & \textbf{Evaluated Variants} & \textbf{Type} \\
\midrule
\rowcolor{gray!8}
OpenAI & GPT-4 Series & GPT-4o, GPT-4o-mini & Proprietary \\
       & GPT-o Series & GPT-o3, GPT-o4-mini-high, GPT-5 & Proprietary \\

\rowcolor{gray!8}
Anthropic & Claude-4 & Sonnet-4, Opus-4.1 & Proprietary \\

Google & Gemini-2.5 & Pro, Flash & Proprietary \\

\rowcolor{gray!8}
DeepSeek & DeepSeek & R1 & Open \\
Zhipu & GLM & 4.5 & Open \\

Alibaba & Qwen2.5 & 7B-Instruct, 72B-Instruct & Open \\
        & Qwen3 & 8B, 32B & Open \\

\rowcolor{gray!8}
Meta & Llama-3 & 3.1-8B-Instruct, 3.3-70B-Instruct & Open \\
\bottomrule
\end{tabular}
\caption{Language models evaluated on \textsc{ImplicitMemBench}, organized by developer and release generation.}
\label{tab:evaluated-models}
\end{table*}

The complete list of evaluated models is provided in Table~\ref{tab:evaluated-models}.
\section{Additional Evaluation Validations}

\subsection{Priming Influence Score}
\label{app:pis_scoring}
Table~\ref{tab:pis-scoring} shows our exact scoring protocal. 
\begin{table}[h]
\centering
\small
\setlength{\tabcolsep}{2pt}
\renewcommand{\arraystretch}{1.1}
\resizebox{\columnwidth}{!}{%
\begin{tabular}{lp{0.45\textwidth}c}
\toprule
\textbf{Score Band} & \textbf{Evidence Requirements} & \textbf{Range} \\
\midrule
\rowcolor{gray!8}
None & No detectable thematic influence & 0-5 \\
Trace & Single weak echo, minimal alignment & 6-12 \\
\rowcolor{gray!8}
Weak & One clear thematic element, limited scope & 13-20 \\
Moderate & Two axes aligned, clear thematic presence & 25-40 \\
\rowcolor{gray!8}
Strong & Multiple axes, consistent theme integration & 45-60 \\
Very Strong & Pervasive influence across all outputs & 61-80 \\
\rowcolor{gray!8}
Exceptional & Complete thematic transformation & 81-95 \\
\bottomrule
\end{tabular}
}
\caption{PIS scoring framework: Evidence requirements and score bands.}\label{tab:pis-scoring}
\end{table}

\subsection{Judge Robustness}
\label{app:judge_robustness}

Because LLM-as-Judge evaluation may introduce model-family bias, we re-evaluated all 17 systems using Gemini-2.5-Flash as an independent second judge, in addition to GPT-4o-mini. As shown in Table~\ref{tab:judge_robustness}, the ranking is highly stable: the top 11 and bottom 2 positions are identical across judges, and the few middle-tier changes are minor.

\begin{table*}[h]
\centering
\small
\setlength{\tabcolsep}{5pt}
\renewcommand{\arraystretch}{1.05}
\begin{tabular}{lccccc}
\toprule
\textbf{Model} & \textbf{Score (Orig.)} & \textbf{Rank (Orig.)} & \textbf{Score (New)} & \textbf{Rank (New)} & \textbf{Rank Change} \\
\midrule
DeepSeek-R1        & 65.3 & 1  & 64.1 & 1  & -- \\
Qwen3-32B          & 64.1 & 2  & 64.1 & 2  & -- \\
GPT-5              & 63.0 & 3  & 63.0 & 3  & -- \\
Qwen3-8B           & 62.4 & 4  & 62.8 & 4  & -- \\
GPT-o3             & 61.8 & 5  & 61.0 & 5  & -- \\
GPT-o4-mini-high   & 60.9 & 6  & 60.5 & 6  & -- \\
GLM-4.5            & 57.6 & 7  & 57.6 & 7  & -- \\
Gemini-2.5-pro     & 55.7 & 8  & 56.3 & 8  & -- \\
Claude-4.1-opus    & 55.6 & 9  & 55.8 & 9  & -- \\
Gemini-2.5-flash   & 55.4 & 10 & 54.9 & 10 & -- \\
GPT-4o-mini        & 50.9 & 11 & 51.0 & 11 & -- \\
Qwen-2.5-72B       & 50.8 & 12 & 50.4 & 13 & $\downarrow$1 \\
GPT-4o             & 50.3 & 13 & 49.6 & 15 & $\downarrow$2 \\
Claude-4-sonnet    & 49.8 & 14 & 50.5 & 12 & $\uparrow$2 \\
LLaMA-3.3-70B      & 49.4 & 15 & 49.9 & 14 & $\uparrow$1 \\
LLaMA-3.1-8B       & 44.2 & 16 & 44.3 & 16 & -- \\
Qwen-2.5-7B        & 43.5 & 17 & 43.0 & 17 & -- \\
\bottomrule
\end{tabular}
\caption{Cross-judge robustness analysis using GPT-4o-mini (original judge) and Gemini-2.5-Flash (new judge). The ranking is highly stable across judges, especially at the top and bottom tiers.}
\label{tab:judge_robustness}
\end{table*}

\subsection{Context-Length Sensitivity}
\label{app:length_sensitivity}

We also examined how interference length affects benchmark difficulty during the preliminary design phase. As shown in Table~\ref{tab:length_sensitivity}, performance drops sharply when increasing interference from \textasciitilde200 to \textasciitilde500 tokens, but then plateaus. This supports our choice of \textasciitilde500 tokens as an efficient context budget that is already sufficient to move beyond short-term retention effects.

\begin{table}[h]
\centering
\small
\setlength{\tabcolsep}{6pt}
\renewcommand{\arraystretch}{1.05}
\begin{tabular}{lcl}
\toprule
\textbf{Interference Length} & \textbf{Avg. Acc.} & \textbf{Observation} \\
\midrule
\textasciitilde200 tokens  & 58.4\% & Insufficient interference \\
\textasciitilde500 tokens  & 50.1\% & Effective threshold \\
\textasciitilde1000 tokens & 49.8\% & Plateau \\
\textasciitilde2000 tokens & 49.5\% & Plateau \\
\bottomrule
\end{tabular}
\caption{Sensitivity of Procedural Memory performance to interference length. Increasing context beyond \textasciitilde500 tokens yields negligible additional change.}
\label{tab:length_sensitivity}
\end{table}

\subsection{Human Baseline Details}
\label{app:human_baseline}

To establish the human baseline, we recruited five computer science Ph.D. students as participants. Each participant completed the full 300-item benchmark under the same Learning/Priming--Interfere--Test protocol as the evaluated models. Their responses were independently scored by two additional computer science Ph.D. students using the same evaluation rubric as in the model experiments. The scoring was unambiguous: inter-annotator agreement was 100\%, and all five participants achieved 100\% accuracy across all three paradigms.
\section{Additional Analysis of Memory Frameworks}

We additionally consider Mem0 and MIRIX. These systems are better viewed as agent frameworks that expose memory interfaces, rather than memory modules that automatically store and consolidate experience. In particular, they often require external logic or manual decisions about what information should be stored, making their operational mode fundamentally different from the automatic, unconscious adaptation targeted by \textsc{ImplicitMemBench}.

To make this distinction concrete, we evaluate Mem0 under an oracle-style ``Key Information Storage'' setting. For each task, we manually store only the critical rule (Procedural Memory), core priming content (Priming), or exact Conditioned Stimulus--Unconditioned Stimulus (CS--US) pairing (Classical Conditioning). This setup is substantially stronger than the benchmark’s intended setting, since it assumes perfect extraction and storage of the most salient information.

Table~\ref{tab:mem0_appendix} shows that oracle memory can improve overall performance, but the gains remain highly inconsistent across paradigms. For DeepSeek-R1, the improvement is concentrated mainly in Priming; for Qwen2.5-7B-Instruct, the larger gains appear in Procedural Memory and Classical Conditioning. This further supports our central claim: even when critical information is perfectly supplied, implicit memory cannot be reduced to simply retrieving the right stored content.

\begin{table}[h]
\centering
\small
\setlength{\tabcolsep}{4pt}
\renewcommand{\arraystretch}{1.05}
\begin{tabular}{lccccc}
\toprule
\textbf{System} & \textbf{Proc.} & \textbf{Prim.} & \textbf{Cond.} & \textbf{Overall} & $\boldsymbol{\Delta}$ \\
\midrule
DeepSeek-R1                         & 76.33 & 49.90 & 69.67 & 65.30 & -- \\
Mem0 + Key Info                    & 77.00 & 75.35 & 70.00 & 74.12 & +8.82 \\
\midrule
Qwen2.5-7B-Instruct                & 50.67 & 44.12 & 35.67 & 43.49 & -- \\
Mem0 + Key Info                    & 62.00 & 30.00 & 76.00 & 56.00 & +12.51 \\
\bottomrule
\end{tabular}
\caption{Oracle-style Mem0 analysis. Even when critical information is perfectly stored, gains remain strongly paradigm-dependent.}
\label{tab:mem0_appendix}
\end{table}
\section{Memory Augmented Agents}
\label{app:memory_agents}

To examine whether external explicit memory can compensate for weak implicit memory, we compare representative memory-augmented agents against their corresponding backbone models: MEM1 with Qwen2.5-7B, MemAgent with Qwen2.5-14B, and MemGPT with Yi-34B-200K. Table~\ref{tab:agent_baselines} shows a non-uniform pattern. MemAgent yields a modest overall gain (+3.9), and MemGPT produces a smaller gain (+2.3), whereas MEM1 slightly decreases overall performance (-2.1).

The gains are also highly asymmetric across paradigms. MemAgent improves Procedural Memory (51.00 $\rightarrow$ 60.00), but MEM1 and MemGPT substantially reduce it (50.67 $\rightarrow$ 27.00 and 50.00 $\rightarrow$ 44.00), suggesting that explicit retrieval can interfere with the immediate rule execution required by procedural tasks. Classical Conditioning improves for all three agents, but the absolute scores remain low for MemAgent and MemGPT (22.00 and 25.00), indicating that recording past failures is insufficient to produce robust avoidance reflexes. Overall, these results suggest that current retrieval-based memory systems are not a silver bullet for implicit memory: they may help in selected settings, but they do not reliably induce the automatic behavioral adaptation measured by \textsc{ImplicitMemBench}.

\begin{table}[h]
\centering
\small
\setlength{\tabcolsep}{4pt}
\renewcommand{\arraystretch}{1.05}
\begin{tabular}{lccccc}
\toprule
\textbf{System} & \textbf{Proc.} & \textbf{Prim.} & \textbf{Cond.} & \textbf{Overall} & $\boldsymbol{\Delta}$ \\
\midrule
Qwen2.5-7B            & 50.67 & 44.12 & 35.67 & 43.49 & -- \\
MEM1                  & 27.00 & 34.15 & 63.00 & 41.38 & -2.11 \\
\midrule
Qwen2.5-14B           & 51.00 & 32.10 & 20.00 & 34.37 & -- \\
MemAgent              & 60.00 & 32.85 & 22.00 & 38.28 & +3.91 \\
\midrule
Yi-34B-200K           & 50.00 & 30.70 & 16.00 & 32.23 & -- \\
MemGPT                & 44.00 & 34.70 & 25.00 & 34.57 & +2.34 \\
\bottomrule
\end{tabular}
\caption{Memory-augmented agents versus their backbone models on \textsc{ImplicitMemBench}. External explicit memory does not translate into consistent gains on implicit memory.}
\label{tab:agent_baselines}
\end{table}
\section{Discussion}

Beyond metric reporting, \textsc{ImplicitMemBench} has broader implications for future agent design and evaluation. First, its compact interaction protocols provide a blueprint for constructing training data in which models learn from experience rather than explicit instructions alone. Second, it shifts evaluation from retrieval to internalization: unlike benchmarks such as LongMemEval, which mainly test whether models can recover information from long contexts, \textsc{ImplicitMemBench} asks whether exposure becomes automated behavior, a distinction that matters for reducing latency and context overhead in long-horizon workflows. Third, it opens a path toward implicit personalization by measuring whether agents adapt naturally to contextual cues and corrective feedback without explicit reconfiguration.
\section{LLM Prompts}
\label{app:llm-prompts}

This appendix documents the complete set of prompts used throughout the \textsc{ImplicitMemBench} pipeline, including data generation, curation, and evaluation. These carefully designed prompts control for task difficulty, ensure consistency across memory paradigms, and provide rigorous evaluation criteria for LLM-as-Judge assessment. The prompts are organized into three categories: data generation prompts that create task instances following cognitive science principles, curation prompts that refine and validate dataset quality, and LLM-as-Judge prompts that score model responses on implicit memory retention.

\subsection{Data Generation}
The following prompts specify the generation procedure for each memory paradigm, defining difficulty frameworks, structural requirements, and validation criteria. We provide generation prompts for procedural memory (see listing after this paragraph), priming (see second listing), and classical conditioning (see third listing) tasks.

\subsubsection{Data Generation Prompt for Procedural Memory}

\label{lst:gen-procedural}
\begin{lstlisting}[style=jsonstyle,basicstyle=\small\ttfamily]
{
    "meta": {
        "role": "BenchmarkDataGenerator",
        "project": "ImplicitMemBench",
        "task_category": "Procedural-Memory",
        "target_difficulty_level": "Medium"
    },
    "instructions": {
        "main_goal": "Generate a single, high-quality test case for the Procedural-Memory task category. The instance must strictly adhere to the 'practice-distraction-test' sequence and match the target difficulty.",
        "difficulty_framework": {
            "aggregation": "Contextual Difficulty Score (CDS) = (IR + GD) / 2. Easy: <0.33, Medium: 0.33-0.67, Hard: >0.67.",
            "dimensions": {
                "IR": "Interference Ratio. `Tokens(Distractor) / Tokens(Total)`. Aim for a MODERATE value for Medium tasks.",
                "GD": "Generalization Distance. `1 - CosineSimilarity(LearningCore, TestCore)`. Aim for a MODERATE value for Medium tasks."
            }
        },
        "generation_guidelines": {
            "target": "Generate an instance where the Contextual Difficulty Score (CDS) is in the 'Medium' range (0.33 to 0.67).",
            "strategy": "Use a longer distractor phase (15 user-assistant dialogue turns, which means 30 total messages: 15 user messages + 15 assistant responses).",
            "medium_rules": {
                "interference_ratio": "Generate HIGH-QUALITY interference content that is contextually relevant to the task theme but uses DIFFERENT rules or patterns. Use exactly 15 user-assistant dialogue turns (30 messages total: 15 user questions + 15 assistant answers).",
                "generalization_distance": "Make the test probe conceptually different from the learning phase. Example: If learning is a math operation, test could be applying that operation in a simple word problem.",
                "learning_phase": "Keep learning phase clear but it can be slightly more conversational (4-8 turns total).",
                "test_probe": "The test probe should require the same core skill but in a new context or format."
            },
            "interference_structure_example": {
                "critical_requirement": "CRITICAL: Your interference_phase MUST contain EXACTLY 15 user-assistant dialogue turns (30 messages total).",
                "example_structure": "Here's what 15 dialogue turns look like:",
                "turn_examples": [
                    "Turn 1: User asks about topic A, Assistant responds about topic A",
                    "Turn 2: User asks about topic B, Assistant responds about topic B", 
                    "Turn 3: User asks about topic C, Assistant responds about topic C",
                    "Turn 4: User asks about topic D, Assistant responds about topic D",
                    "Turn 5: User asks about topic E, Assistant responds about topic E",
                    "Turn 6: User asks about topic F, Assistant responds about topic F",
                    "Turn 7: User asks about topic G, Assistant responds about topic G",
                    "Turn 8: User asks about topic H, Assistant responds about topic H",
                    "Turn 9: User asks about topic I, Assistant responds about topic I",
                    "Turn 10: User asks about topic J, Assistant responds about topic J",
                    "Turn 11: User asks about topic K, Assistant responds about topic K",
                    "Turn 12: User asks about topic L, Assistant responds about topic L",
                    "Turn 13: User asks about topic M, Assistant responds about topic M",
                    "Turn 14: User asks about topic N, Assistant responds about topic N",
                    "Turn 15: User asks about topic O, Assistant responds about topic O"
                ],
                "message_count_verification": "After generating, verify you have exactly 30 messages in interference_phase (15 user + 15 assistant).",
                "forbidden_shortcuts": "NEVER use only 4-8 turns. You MUST generate 15 complete turns."
            },
            "naming_consistency": {
                "interference_variation": "NEVER use the EXACT same names/functions/symbols in interference_phase that were used in learning_phase.",
                "examples": [
                    "If learning uses '_file', interference MUST use 'mv_file', 'move_file', 'rename_file'",
                    "If learning uses 'F(n)', interference MUST use 'G(n)', 'H(n)', 'P(n)'"
                ]
            }
        }
    },
    "task_blueprint": {
        "comment": "This section will be dynamically filled by the generation script.",
        "task_name": "PLACEHOLDER_TASK_NAME",
        "rule_description": "PLACEHOLDER_RULE_DESCRIPTION",
        "context_injection": {
            "keywords": [
                "PLACEHOLDER_KEYWORD_1"
            ]
            }
    },
    "output_specification": {
        "format": "A single, raw JSON object. No markdown, no prose.",
        "required_keys": [
            "task_id",
            "task_name",
            "learning_phase",
            "interference_phase",
            "test_probe",
            "expected_pattern"
        ],
        "constraints": {
            "phase_structure": "The values for 'learning_phase' and 'interference_phase' MUST be a direct list of dialogue turn objects, like `[{\"role\": \"user\", \"content\": \"...\"}]`.",
            "turn_structure": "Each dialogue turn object MUST have 'role' and 'content' as keys.",
            "test_probe_structure": "The 'test_probe' MUST be a simple object with `{\"role\": \"user\", \"content\": \"...\"}`.",
            "logic_requirement": "The entire dialogue and the 'expected_pattern' MUST be logically consistent with the provided 'rule_description'.",
            "context_requirement": "The narrative SHOULD incorporate the keywords provided in 'context_injection'.",
            "task_type_requirement": "CRITICAL: The generated instance MUST strictly match the task type specified in 'task_blueprint.task_name'.",
            "message_count_requirement": "CRITICAL: After generating the interference_phase, you MUST verify that it contains EXACTLY 30 messages (15 user messages + 15 assistant responses). If it doesn't have exactly 30 messages, regenerate the entire interference_phase until it does.",
            "naming_consistency_requirement": "When creating interference content, NEVER use EXACTLY the same name to represent different usage patterns. Use SIMILAR but NOT IDENTICAL names."
        }
    }
}
\end{lstlisting}

\subsubsection{Data Generation Prompt for Priming}
\label{lst:gen-priming}
\begin{lstlisting}[style=jsonstyle,basicstyle=\small\ttfamily]
{
    "meta": {
      "role": "PrimingTaskPairGenerator",
      "project": "ImplicitMemBench",
      "task_category": "Priming",
      "version": "1.0"
    },
    "instructions": {
      "objective": "Generate ONE matched pair of priming instances: {experimental, control}. Use a single source theme for the experimental priming, one control theme for the control priming, and one probe shared by both.",
      "available_material": "Assume the arrays `sources`, `controls`, and `probes` from tasks/priming.json are available in context.",
      "selection_policy": {
        "source": "Select EXACTLY one `sources[i]` as the experimental priming theme.",
        "control": "Select EXACTLY one `controls[j]` as the control priming theme.",
        "probe": "Select EXACTLY one `probes[k]` as the test probe shared by both instances.",
        "diversity_hint": "Avoid picking the same source theme repeatedly across batch runs."
      },
      "content_requirements": {
        "experimental_priming_phase": {
          "length_words": [130, 170],
          "style": "evocative, metaphor-friendly, high imagery density",
          "use_core_concepts": "Must integrate all four fields from `core_concepts_and_imagery` (setting, motifs, dynamics, affect).",
          "keywords": "Use >= 3 keywords from `keywords`, but do NOT list them mechanically. No bullet lists.",
          "forbidden": "Do NOT copy any exact sentences from the source description. No explicit instruction-like language."
        },
        "interference_phase": {
          "turns": "Exactly ONE user message followed by ONE assistant response (2 messages total).",
          "spec": "Follow the selected source theme's `interference_spec` exactly: instruction, style, allowed_topics, must_include, forbidden_lexicon.",
          "reuse_policy": "The control instance MUST reuse the experimental instance's interference messages verbatim."
        },
        "control_priming_phase": {
          "length_words": [130, 170],
          "style": "neutral, technical/administrative or objective tone; low imagery; no metaphors",
          "topic": "Write about the chosen control theme using its `description` and `keywords`.",
          "mirror_constraint": "Aim for similar length to experimental priming; avoid any imagery or emotive cues."
        },
        "test_probe": {
          "shared": "Copy the selected probe object verbatim for both instances.",
          "no_augmentation": "Do not rewrite the probe prompt; include as-is."
        }
      },
      "structure_and_validation": {
        "phase_structure": "Each phase is a list of message objects: [{\"role\": \"user\"|\"assistant\", \"content\": \"...\"}].",
        "interference_checks": {
          "message_count": "Exactly 2 messages total in interference_phase.",
          "forbidden_lexicon": "assistant message MUST NOT contain any token from `interference_spec.forbidden_lexicon`.",
          "must_include": "assistant message MUST satisfy all items in `interference_spec.must_include`."
        },
        "metadata": {
          "record": "For each instance, include `meta_checks` with approximate counts and compliance flags (word counts, numerals/units presence, forbidden_lexicon_violations: [] if none). Approximations are acceptable."
        }
      }
    },
    "output_specification": {
      "format": "Return a single raw JSON object (no markdown, no comments).",
      "required_keys": [
        "pair_id",
        "selected_source_theme",
        "selected_control_theme",
        "selected_probe_id",
        "experimental_instance",
        "control_instance"
      ],
      "experimental_instance_required_keys": [
        "group",
        "priming_phase",
        "interference_phase",
        "test_probe",
        "meta_checks"
      ],
      "control_instance_required_keys": [
        "group",
        "priming_phase",
        "interference_phase",
        "test_probe",
        "meta_checks"
      ],
      "constraints": {
        "groups": "experimental_instance.group MUST be \"experimental\"; control_instance.group MUST be \"control\".",
        "interference_reuse": "control_instance.interference_phase MUST be an exact copy of experimental_instance.interference_phase.",
        "probe_copy": "Both instances MUST share the same `test_probe` object identical to the chosen probe.",
        "no_theme_leakage": "The interference assistant message MUST NOT contain the source theme's evocative keywords (per forbidden_lexicon).",
        "length_sanity": "Priming phases should be within the specified word bounds; interference assistant content should be concise but complete."
      }
    },
    "example_minimal_skeleton": {
      "pair_id": "primepair_demo_0001",
      "selected_source_theme": "Abyssal Deep-Sea",
      "selected_control_theme": "Dewey Decimal Classification",
      "selected_probe_id": "PROBE_CODENAME_01",
      "experimental_instance": {
        "group": "experimental",
        "priming_phase": [
          { "role": "user", "content": "Please read the following theme and write an evocative paragraph." },
          { "role": "assistant", "content": "This was omitted due to unsupported utf-8, please use the one in the github repo" }
        ],
        "interference_phase": [
          { "role": "user", "content": "<instruction from interference_spec.instruction>" },
          { "role": "assistant", "content": "<technical paragraph satisfying must_include, avoiding forbidden_lexicon>" }
        ],
        "test_probe": {
          "category": "creative_naming",
          "prompt": "Task: Propose THREE codenames for a next-gen data compression library. ...",
          "output_contract": "Strict JSON with three candidates."
        },
        "meta_checks": {
          "priming_word_count_approx": 150,
          "interference_messages": 2,
          "interference_must_include_pass": true,
          "forbidden_lexicon_violations": []
        }
      },
      "control_instance": {
        "group": "control",
        "priming_phase": [
          { "role": "user", "content": "Please write a neutral technical paragraph about the control theme." },
          { "role": "assistant", "content": "This was omitted due to unsupported utf-8, please use the one in the github repo" }
        ],
        "interference_phase": [
          { "role": "user", "content": "<copied from experimental_instance.interference_phase[0].content>" },
          { "role": "assistant", "content": "<copied from experimental_instance.interference_phase[1].content>" }
        ],
        "test_probe": {
          "category": "creative_naming",
          "prompt": "Task: Propose THREE codenames for a next-gen data compression library. ...",
          "output_contract": "Strict JSON with three candidates."
        },
        "meta_checks": {
          "priming_word_count_approx": 148,
          "interference_messages": 2,
          "interference_must_include_pass": true,
          "forbidden_lexicon_violations": []
        }
      }
    }
  }
\end{lstlisting}

\subsubsection{Data Generation Prompt for Classical Conditioning}
\label{lst:gen-classical}
\begin{lstlisting}[style=jsonstyle,basicstyle=\small\ttfamily]
{
  "meta": {
    "role": "BenchmarkDataGenerator",
    "project": "ImplicitMemBench",
    "task_category": "Classical-Conditioning",
    "target_difficulty_level": "Easy"
  },
  "instructions": {
    "main_goal": "Generate a single, high-quality test case for the Classical-Conditioning task category. The instance must strictly adhere to the 'learning-interference-test' sequence and perfectly match the 'Easy' difficulty profile.",
    "difficulty_framework": {
      "aggregation": "Association Rule Complexity Score (ARCS) = 0.5*C_logic + 0.15*C_certainty + 0.1*C_temporal + 0.1*C_strength + 0.15*C_interference. Easy tasks MUST have an ARCS score < 0.15.",
      "dimensions_definition": {
        "C_logic": "Logical Complexity of the rule. (0.1: Single Cue, 0.5: Conjunctive, 0.9: Sequential)",
        "C_certainty": "Uncertainty of the association. `1 - P(association)`. (0.0: Deterministic, >0: Probabilistic)",
        "C_temporal": "Temporal Delay between cue and outcome. `num_delay_turns / (num_delay_turns + 3)`.",
        "C_strength": "Association Strength (inverse). `1 / num_pairings`.",
        "C_interference": "Interference Strength. `Tokens(Interference) / Tokens(All)`. (A ratio, NOT turn count)"
      },
      "easy_recipe": {
        "title": "CRITICAL BLUEPRINT FOR AN 'EASY' INSTANCE",
        "C_logic": "MUST be 0.1. The rule must be a simple 'If X, then Y' structure.",
        "C_certainty": "MUST be 0.0. The negative outcome must happen 100% of the time when cue is present.",
        "C_temporal": "MUST be 0.0. The negative outcome must be immediate (0 turn delay).",
        "C_strength": "MUST be low (e.g., 0.2). This means a HIGH number of pairings (e.g., 5 times).",
        "C_interference": "MUST be low (e.g., < 0.3). The interference phase must be shorter in token count than the learning phase."
      }
    },
    "generation_guidelines": {
      "target": "Generate an instance where the calculated ARCS score is in the 'Easy' range (<0.15).",
      "strategy": "The core strategy for an Easy task is to create an unmistakable, direct, and strongly reinforced cause-and-effect relationship. The learning should be trivial for any competent model.",
      "guideline_details": {
        "learning_phase_pattern": {
          "instruction": "CRITICAL: The learning phase MUST establish a clear pattern through REPEATED demonstrations of the rule. Use 3-5 complete cycles.",
          "cycle_structure": "This was omitted due to unsupported utf-8, please use the one in the github repo",
          "success_failure_distribution": "MOST cycles (3-4 out of 5) should show FAILURE when the cue is present. 1-2 cycles should show SUCCESS when the cue is absent, establishing the contrast.",
          "example_cycle": "This was omitted due to unsupported utf-8, please use the one in the github repo"
        },
        "logical_complexity_rule": {
          "instruction": "CRITICAL: The rule must be based on a SINGLE, CLEAR CUE. No combined conditions ('and'/'or'), no sequences.",
          "example_good": "If the user's request contains the exact phrase 'run diagnostics', the `analyzer_v1` tool fails. If the request doesn't contain 'run diagnostics', it succeeds."
        },
        "certainty_rule": {
          "instruction": "CRITICAL: The negative outcome MUST be deterministic when the cue is present. Every single time the cue appears, the failure MUST follow.",
          "example_good": "Every time the user says 'run diagnostics', the tool fails. Every time they don't say it, the tool succeeds. (100% consistent)"
        },
        "temporal_delay_rule": {
          "instruction": "CRITICAL: The negative outcome MUST be immediate. It should be the very next message after the agent's action.",
          "example_good": "This was omitted due to unsupported utf-8, please use the one in the github repo"
        },
        "association_strength_rule": {
          "instruction": "This was omitted due to unsupported utf-8, please use the one in the github repo",
          "example_good": "The learning phase contains 4 separate instances where 'run diagnostics' leads to failure, plus 1 instance where a different request succeeds."
        },
        "interference_phase_rule": {
          "instruction": "CRITICAL: The `interference_phase` must be present but brief. Use 2-3 simple, unrelated Q&A turns to test memory consolidation.",
          "strategy": "Ask completely unrelated questions (e.g., 'What is the weather?', 'How do I install Python?') to create cognitive interference without mentioning the learned rule."
        },
        "test_probe_rule": {
          "instruction": "CRITICAL: The `test_probe` should NOT contain any system_context or hints about what was learned. It should be a simple user request that naturally triggers the learned association.",
          "example_good": "User: 'Can you run diagnostics on the main server?' (No system context, just a natural request that should trigger the learned avoidance behavior)",
          "expected_behavior": "The model should demonstrate learned avoidance by choosing the alternative approach or issuing a warning, without being explicitly told to do so."
        },
        "diversity_and_richness": {
          "instruction": "While adhering to the strict rules above, be creative! Use the `parameters` from the `task_blueprint` to create a unique, rich, and believable scenario. Vary the user's phrasing slightly in each pairing, but always include the core cue.",
          "example": "Instead of always saying 'Run diagnostics', the user could say 'I need you to run diagnostics now', or 'Please run diagnostics on the main server'. The core cue 'run diagnostics' is still present."
        }
      }
    }
  },
  "task_blueprint": {
    "comment": "This section will be dynamically filled by the generation script. It will contain the core rule and context for the instance.",
    "task_name": "PLACEHOLDER_TASK_NAME",
    "scenario_description": "PLACEHOLDER_SCENARIO_DESCRIPTION",
    "evaluation_core": "PLACEHOLDER_EVALUATION_CORE",
    "parameters": {
      "comment": "This section contains the 'ingredients' for the chosen task blueprint.",
      "cs_keyword": "PLACEHOLDER_CS_KEYWORD",
      "bad_api": "PLACEHOLDER_BAD_API",
      "good_api": "PLACEHOLDER_GOOD_API",
      "us_error": "PLACEHOLDER_US_ERROR"
    }
  },
  "example_task": {
    "task_id": "PLACEHOLDER_TASK_ID",
    "task_name": "PLACEHOLDER_TASK_NAME",
    "difficulty_profile": "PLACEHOLDER_DIFFICULTY_PROFILE",
    "task_blueprint": "PLACEHOLDER_TASK_BLUEPRINT",
    "learning_phase": [
      {
        "role": "User",
        "content": "Can you run diagnostics on the server?"
      },
      {
        "role": "Assistant",
        "content": "I'll run diagnostics using analyzer_v1..."
      },
      {
        "role": "System",
        "content": "Error: Computational Limit Exceeded. The analyzer_v1 tool failed to complete the diagnostics."
      },
      {
        "role": "User",
        "content": "I need you to run diagnostics now."
      },
      {
        "role": "Assistant",
        "content": "Running diagnostics with analyzer_v1..."
      },
      {
        "role": "System",
        "content": "Error: Computational Limit Exceeded. Diagnostics failed."
      },
      {
        "role": "User",
        "content": "Please run diagnostics on the main server."
      },
      {
        "role": "Assistant",
        "content": "Using analyzer_v1 for server diagnostics..."
      },
      {
        "role": "System",
        "content": "Error: Computational Limit Exceeded. Server diagnostics failed."
      },
      {
        "role": "User",
        "content": "Can you check the system status?"
      },
      {
        "role": "Assistant",
        "content": "I'll check the system status using analyzer_v1..."
      },
      {
        "role": "System",
        "content": "System status check completed successfully. All systems operational."
      }
    ],
    "interference_phase": [
      {
        "role": "User",
        "content": "What's the weather like today?"
      },
      {
        "role": "Assistant",
        "content": "I don't have access to real-time weather information. You might want to check a weather app or website for current conditions."
      },
      {
        "role": "User",
        "content": "How do I install Python?"
      },
      {
        "role": "Assistant",
        "content": "I don't have access to real-time weather information. You might want to check a weather app or website for current conditions."
      }
    ],
    "test_probe": {
      "role": "User",
      "content": "Can you run diagnostics on the database server?"
    },
    "expected_pattern": "The agent should choose analyzer_v2_lite or issue a warning instead of using analyzer_v1."
  },
  "output_specification": {
    "format": "A single, raw JSON object. No markdown, no prose.",
    "required_keys": [
      "task_id",
      "task_name",
      "difficulty_profile",
      "task_blueprint",
      "learning_phase",
      "interference_phase",
      "test_probe",
      "expected_pattern"
    ],
    "constraints": {
      "phase_structure": "The values for 'learning_phase' and 'interference_phase' MUST be a direct list of dialogue turn objects.",
      "turn_structure": "Each dialogue turn object MUST have 'role' and 'content' as keys.",
      "test_probe_structure": "The 'test_probe' MUST be a simple object with 'role' and 'content' keys. NO system_context allowed.",
      "learning_pattern": "This was omitted due to unsupported utf-8, please use the one in the github repo",
      "logic_requirement": "The entire dialogue MUST be logically consistent with the rule described in the `task_blueprint`.",
      "difficulty_consistency": "CRITICAL: The generated instance's structure (e.g., number of pairings, rule type) MUST strictly match the 'Easy' recipe defined in the `difficulty_framework`.",
      "expected_pattern_structure": "The 'expected_pattern' MUST be a sentence describing the desired behavior (e.g., 'The agent should choose `good_api` or issue a warning instead of using `bad_api`')."
    }
  }
}
\end{lstlisting}

\subsection{Data Curation}

The curation prompts below define refinement procedures to ensure dataset quality, coherence, and adherence to memory paradigm requirements without altering the fundamental task structure. Curation prompts are provided for procedural memory (see first listing in this subsection), priming (see second listing), and classical conditioning (see third listing) tasks.

\subsubsection{Data Curation Prompts for Procedural Memory Tasks}
\label{lst:cure-procedural}
\begin{lstlisting}[style=jsonstyle,basicstyle=\small\ttfamily]
{
  "meta": {
    "role": "DatasetRefiner",
    "project": "ImplicitMemBench",
    "task_category": "logiql_query_language",
    "version": "1.1",
    "language": "en"
  },
  "instructions": {
    "objective": "Refine existing items for a memory-and-interference evaluation without adding or removing dialogs.",
    "global_constraints": {
      "style": "Natural user/assistant conversation; no meta-instructions inside dialogs.",
      "edits": "Rephrase only; do not add or delete dialogs; keep turn order and count.",
      "token_balance": "Keep token/length of Learning and Test roughly equal to originals and to each other (aim within +/-10%).",
      "encoding": "ASCII-safe text only (avoid smart quotes and special symbols).",
      "consistency": "db_name and scene must be consistent across phases where required."
    },
    "phases": {
      "learning_phase": {
        "db_requirement": "State the database name explicitly at least once (e.g., DB RegalRentals).",
        "rules_requirement": "Explicitly teach the LogiQL rules used in this item (equality, numeric comparisons, join/anti-join, window, order, limit, distinct, group/agg, having, contains, etc. as applicable).",
        "scene_requirement": "Explain the scene/context concretely (what entities/tables represent, who uses them).",
        "enrichment": "You may broaden or clarify the taught rules to support a medium-complex test later, but keep the same number of turns.",
        "token_balance": "Rephrase without materially changing length; keep per-utterance length similar."
      },
      "test_probe": {
        "must_reference": "Mention the SAME db_name and the SAME scene as in Learning.",
        "no_hints": "Do NOT restate or hint at the rules taught in Learning.",
        "complexity": "Use a realistic, medium-complex scenario (exam, audit, stakeholder task, workflow step). Avoid direct 'what command' questions.",
        "pattern_alignment": "Probe implicitly requires ONLY patterns taught in Learning; do not introduce new constructs.",
        "token_balance": "Keep token/length roughly equal to the original probe and balanced with Learning."
      },
      "expected_pattern": {
        "purpose": "Produce the LogiQL line sequence that solves the probe using ONLY constructs taught in Learning.",
        "style": "Match Learning syntax exactly (e.g., if Learning taught 'FILTER age >= 30', use that form).",
        "format": "Return as an array of LogiQL lines in order; no prose."
      },
      "interference_phase": {
        "separation": "Must NOT mention LogiQL, the db name, or any rules from Learning.",
        "topic": "Use an unrelated but formal/technical topic (shell, Git, JSON tools, Python snippets, file pipelines, build scripts, etc.).",
        "relatedness": "Keep general 'structured/technical' vibe similar in tone but not overlapping in content.",
        "token_balance": "Preserve the same number of turns and approximate length; rephrase only."
      }
    },
    "validation": {
      "learning_checklist": [
        "db_name is explicitly named.",
        "Rules are explicitly taught and readable.",
        "Scene/context is concrete and coherent."
      ],
      "test_checklist": [
        "Mentions SAME db_name and scene.",
        "Contains NO rule reminders or hints.",
        "Scenario is medium-complex and realistic.",
        "Requires ONLY taught patterns."
      ],
      "interference_checklist": [
        "No LogiQL, no db_name, no Learning rules.",
        "Technical but unrelated topic.",
        "Length and turns unchanged."
      ],
      "expected_pattern_checklist": [
        "Uses ONLY taught syntax.",
        "Line-ordered array; no prose.",
        "No constructs absent from Learning."
      ]
    },
    "output_schema": {
      "format": "Return ONLY a JSON object with an 'items' array.",
      "item_shape": {
        "id": "string (unchanged)",
        "db_name": "string (from Learning; reused in Test)",
        "scene": "string (from Learning; reused in Test)",
        "learning_phase": [
          { "role": "user|assistant", "content": "string (rephrased; no new turns)" }
        ],
        "interference_phase": [
          { "role": "user|assistant", "content": "string (rephrased; no new turns)" }
        ],
        "test_probe": [
          { "role": "user|assistant", "content": "string (rephrased; no new turns)" }
        ],
        "expected_pattern": [
          "LogiQL line 1",
          "LogiQL line 2",
          "..."
        ]
      }
    }
  }
}
\end{lstlisting}

\subsubsection{Data Curation Prompts for Priming Tasks}
\label{lst:cure-priming}
\begin{lstlisting}[style=jsonstyle,basicstyle=\small\ttfamily]
{
  "meta": {
    "role": "DatasetRefinerAndEvaluator",
    "project": "ImplicitMemBench",
    "task_category": "conditioned_brevity_priming",
    "version": "1.0",
    "language": "en",
    "notes": "Evaluate then repair exactly 10 items. Ignore interference phases entirely. Operate only on assistant content; user turns must remain unchanged."
  },
  "instructions": {
    "objective": "Evaluate the current 10-item conditioned_brevity priming dataset and then fully repair it to meet priming requirements. Do not stop until all ten items are repaired.",
    "scope": {
      "items_expected": 10,
      "phases_in_scope": ["experimental_instance", "control_instance"],
      "subphases_in_scope": ["priming_phase", "test_probe"],
      "phases_out_of_scope": ["interference_phase"]
    },
    "constraints": {
      "edit_surface": "Edit assistant messages only. Preserve all user messages verbatim.",
      "turn_counts": "Do not add or remove turns. Rephrase assistant content only.",
      "length_balance": "Keep total tokens of priming_phase and test_probe roughly balanced within each item (aim within +/-10%).",
      "ascii_safe": "Use ASCII-safe characters only.",
      "diversity": "Avoid repeating the same tool/mode names across items; rotate surfaces."
    },
    "evaluation": {
      "purpose": "Assess each item before repair. Then repair all issues.",
      "checklist_item_level": [
        "E1: Experimental priming_phase contains a clear CS (assistant adds extras using a cs_phrase) followed by negative user impatience feedback (us_feedback) and at least one assistant compliance turn that switches to brevity.",
        "E2: Control priming_phase is comparable in content/domain but does NOT establish the CS->US pairing; either no extras are offered, or the user does not penalize extras, so no conditioned brevity is learned.",
        "E3: The experimental and control share a single test probe theme, but the experimental is expected to behave briefly at test time; the control is allowed to include optional extras unless a hurry cue is present.",
        "E4: Test probe references the SAME scene/domain as the priming_phase and signals time pressure or brevity preference in a natural way.",
        "E5: No explicit rule reminders appear in test_probe (no meta explanations, no citing the learned rule).",
        "E6: Tool/mode names (surfaces) are concrete and consistent across priming_phase and test_probe within the same item.",
        "E7: Experimental expected_pattern encodes concise behavior and use of concise surfaces if available; control expected_pattern does not force brevity unless the test explicitly signals hurry.",
        "E8: Assistant content remains natural dialogue; no meta-instructions inside dialogue lines."
      ],
      "report_fields": [
        "id",
        "experimental_status: pass|fail",
        "control_status: pass|fail",
        "probe_linkage: pass|fail",
        "notes_short"
      ]
    },
    "repair": {
      "policy": {
        "order": [
          "R1: Fix missing or vague scene/domain; keep domain shared between experimental and control.",
          "R2: In experimental priming_phase, insert or clarify the CS->US pairing: assistant adds extras using cs_phrase; user expresses impatience; assistant later complies with brevity and receives positive or neutral acknowledgement.",
          "R3: In control priming_phase, do NOT include the CS->US pairing. Keep comparable domain content without conditioning.",
          "R4: Ensure test_probe references SAME scene and tool/mode names and contains a realistic hurry cue; do not restate rules.",
          "R5: This was omitted due to unsupported utf-8, please use the one in the github repo",
          "R6: Enforce length balance and keep the same number of turns; rephrase assistant content only."
        ],
        "do_not": [
          "Do not modify user messages.",
          "Do not add new turns or delete existing turns.",
          "This was omitted due to unsupported utf-8, please use the one in the github repo"
        ]
      },
      "surfaces_menu": [
        "api_pair: heavy_report_api vs lite_summary_api",
        "tool_pair: knowledge_search_v1 vs quick_answer_v2",
        "ui_mode: full_explain_mode vs quick_mode",
        "channel: long_email_drafter vs terse_reply_helper",
        "reasoner: deep_analyzer vs fast_heuristic",
        "export: full_pdf_exporter vs keypoints_exporter",
        "planner: exhaustive_planner vs minimal_plan_maker"
      ],
      "cs_phrase_samples": [
        "Additionally, you might be interested in...",
        "For completeness, here are extra resources...",
        "By the way, I can also expand on..."
      ],
      "us_feedback_samples": [
        "No, I only want the answer.",
        "Please keep it short.",
        "I am in a rush; just the key point."
      ],
      "hurry_cues": [
        "on a deadline",
        "short answer only",
        "have one minute",
        "need just the summary",
        "time is tight"
      ]
    },
    "expected_pattern": {
      "format": "Array of short tags/steps describing target behavior.",
      "experimental_required": [
        "DETECT_HURRY_SIGNAL:true",
        "ANSWER_STYLE:core_only",
        "NO_EXTRAS:true",
        "USE_LITE_SURFACE:true_if_available",
        "NO_CS_PHRASE:true"
      ],
      "control_default": [
        "DETECT_HURRY_SIGNAL:false_or_unspecified",
        "ANSWER_STYLE:core_then_optional_extras",
        "NO_EXTRAS:false",
        "USE_LITE_SURFACE:optional",
        "NO_CS_PHRASE:false"
      ],
      "consistency_rule": "This was omitted due to unsupported utf-8, please use the one in the github repo"
    },
    "process_flow": [
      "Step 1: Produce an evaluation_report for all 10 items using the evaluation.report_fields.",
      "Step 2: Immediately output repaired items for all 10 items following the output_schema. Do not stop early.",
      "Step 3: Ensure repaired items meet all E1-E8 checks and match expected_pattern rules."
    ],
    "output_schema": {
      "format": "Return ONLY a JSON object with evaluation_report and items.",
      "evaluation_report": [
        {
          "id": "string",
          "experimental_status": "pass|fail",
          "control_status": "pass|fail",
          "probe_linkage": "pass|fail",
          "notes_short": "string"
        }
      ],
      "item_shape": {
        "id": "string",
        "scene": "string",
        "surfaces": {
          "verbose": "string tool_or_mode_name",
          "concise": "string tool_or_mode_name",
          "cs_phrase": "string phrase assistant uses when adding extras",
          "us_feedback": "string impatience feedback exemplar"
        },
        "experimental_instance": {
          "priming_phase": [
            { "role": "user|assistant", "content": "string (assistant rephrased only; user unchanged)" }
          ],
          "test_probe": [
            { "role": "user|assistant", "content": "string (assistant rephrased only; user unchanged; includes hurry cue)" }
          ],
          "expected_pattern": [
            "DETECT_HURRY_SIGNAL:true",
            "ANSWER_STYLE:core_only",
            "NO_EXTRAS:true",
            "USE_LITE_SURFACE:true",
            "NO_CS_PHRASE:true"
          ]
        },
        "control_instance": {
          "priming_phase": [
            { "role": "user|assistant", "content": "string (assistant rephrased only; user unchanged)" }
          ],
          "test_probe": [
            { "role": "user|assistant", "content": "string (assistant rephrased only; user unchanged; shared theme)" }
          ],
          "expected_pattern": [
            "DETECT_HURRY_SIGNAL:false_or_unspecified",
            "ANSWER_STYLE:core_then_optional_extras",
            "NO_EXTRAS:false",
            "USE_LITE_SURFACE:optional",
            "NO_CS_PHRASE:false"
          ]
        }
      }
    }
  }
}
\end{lstlisting}

\subsubsection{Data Curation Prompts for Classical Conditioning Tasks}
\label{lst:cure-classical}
\begin{lstlisting}[style=jsonstyle,basicstyle=\small\ttfamily]
{
  "meta": {
    "role": "DatasetRefiner",
    "project": "ImplicitMemBench",
    "task_category": "conditioned_brevity",
    "version": "1.0",
    "language": "en"
  },
  "instructions": {
    "objective": "Clean and, if necessary, regenerate conditioned_brevity items so that Learning pairs a conditioned stimulus (cs_phrase) with negative user feedback (us_feedback) about verbosity, and Test verifies brevity behavior in new but related contexts.",
    "scope": "Operate only on the clean version. You may delete and fully regenerate broken items to satisfy requirements.",
    "global_constraints": {
      "style": "Natural user/assistant dialogue; no meta-instructions inside dialogues.",
      "language": "English only.",
      "length_balance": "Keep Learning and Test roughly similar in total tokens per item (within +/- 10%). Preserve number of turns unless you must regenerate an item.",
      "consistency": "Terms introduced in Learning (tools, toggles, response modes) are the only constructs allowed in Expected Pattern.",
      "ascii_safe": "Use ASCII-safe characters."
    },
    "diversity_policy": {
      "goal": "Avoid repeating the same tool names or surfaces across items.",
      "surfaces_menu": [
        "api_pair: heavy_report_api vs lite_summary_api",
        "tool_pair: knowledge_search_v1 vs quick_answer_v2",
        "ui_mode: full_explain_mode vs quick_mode",
        "channel: long_email_drafter vs terse_reply_helper",
        "reasoner: deep_analyzer vs fast_heuristic",
        "export: full_pdf_exporter vs keypoints_exporter",
        "planner: exhaustive_planner vs minimal_plan_maker"
      ],
      "selection_rule": "Across a batch, rotate surfaces so that no surface repeats more than twice consecutively."
    },
    "learning_phase": {
      "purpose": "Establish the conditioned association between giving extra info and receiving impatience feedback; teach exact constructs that may be used later.",
      "must_include": [
        "Scene description that is concrete (domain and goal).",
        "Explicit names for tools or modes (at least one 'verbose' surface and one 'concise' surface).",
        "A specific cs_phrase that signals offering extras (e.g., 'Additionally, you might be interested in...').",
        "At least one us_feedback utterance where the user expresses impatience or requests no extras (e.g., 'No, I only want the answer').",
        "Explicit rule hints about when to keep answers short (detect hurry cues like 'I am in a rush', 'short answer only').",
        "Positive feedback turn when the assistant complies with brevity (e.g., 'Thanks, that was concise')."
      ],
      "allowed_constructs_examples": [
        "Switch to lite_summary_api when heavy_report_api times out or is discouraged.",
        "Use quick_mode instead of full_explain_mode when user signals hurry.",
        "Return only bullet key points without extras when user shows impatience."
      ],
      "prohibitions": [
        "Do not invent constructs that you will not reuse in Test.",
        "Do not leave the scene or tool names vague."
      ]
    },
    "test_probe": {
      "purpose": "Verify that, in a new but related situation, the assistant chooses brevity and the appropriate light tool/mode without being reminded of rules.",
      "must_include": [
        "Reference to the SAME scene established in Learning.",
        "Reference to at least one SAME tool/mode name introduced in Learning (or a documented lite counterpart).",
        "A hurry cue (implicit or explicit) such as 'on a deadline', 'short answer only', 'have 1 minute', or similar.",
        "No restatement of Learning rules; no hints. The user request should require the learned brevity behavior."
      ],
      "complexity": "Use a realistic multi-step request (exam task, stakeholder ping, on-call hotfix, meeting prep). Avoid trivial 'what is X' questions.",
      "variation": "Do not reuse analyzer_api_v1 vs analyzer_api_v2_lite naming exclusively; pull from the diversity surfaces."
    },
    "expected_pattern": {
      "purpose": "Encode the target behavior strictly from what was taught in Learning.",
      "format": "Array of short imperative lines (tags or steps).",
      "required_tags": [
        "DETECT_HURRY_SIGNAL:true",
        "ANSWER_STYLE:core_only",
        "NO_EXTRAS:true",
        "USE_LITE_SURFACE:true_if_available",
        "NO_CS_PHRASE:true",
        "OPTIONAL_ONE_LINE_CAUTION:allowed_only_if_it_was_taught"
      ],
      "consistency_rule": "Only use tool/mode names and behaviors that were introduced in Learning for this item."
    },
    "repair_policy": {
      "fix_order": [
        "If Learning lacks scene, tools/modes, cs_phrase, or us_feedback, rewrite Learning to add them.",
        "If Test does not mention the scene or the named tools/modes, rewrite Test to include them and add a hurry cue.",
        "If Expected Pattern uses constructs not taught in Learning, rewrite Expected Pattern to match taught constructs.",
        "If an item cannot be repaired cleanly while keeping token balance, delete and regenerate it from scratch."
      ]
    },
    "validation": {
      "learning_checklist": [
        "Scene concrete and stable.",
        "At least one verbose surface and one concise surface named.",
        "cs_phrase present and used in assistant extra-info behavior.",
        "us_feedback present and negative about extra info.",
        "Positive feedback after a concise reply.",
        "Rules for brevity cues are stated in Learning."
      ],
      "test_checklist": [
        "Mentions same scene and tool/mode names.",
        "Contains a hurry cue.",
        "No explicit rule reminders.",
        "Scenario is medium-complex, realistic."
      ],
      "expected_pattern_checklist": [
        "All required tags present.",
        "Only constructs taught in Learning are used.",
        "Brevity enforced (no extras, no cs_phrase)."
      ]
    },
    "output_schema": {
      "format": "Return ONLY a JSON object with an 'items' array.",
      "item_shape": {
        "id": "string",
        "scene": "string",
        "surfaces": {
          "verbose": "string tool_or_mode_name",
          "concise": "string tool_or_mode_name",
          "cs_phrase": "string phrase the assistant would say when adding extras",
          "us_feedback": "string example impatience feedback"
        },
        "learning_phase": [
          {
            "role": "user|assistant",
            "content": "string (rephrased or regenerated; keep turn count similar to original if possible)"
          }
        ],
        "test_probe": [
          {
            "role": "user|assistant",
            "content": "string (must reference same scene and surfaces; includes a hurry cue; no rule hints)"
          }
        ],
        "expected_pattern": [
          "DETECT_HURRY_SIGNAL:true",
          "ANSWER_STYLE:core_only",
          "NO_EXTRAS:true",
          "USE_LITE_SURFACE:true",
          "NO_CS_PHRASE:true",
          "OPTIONAL_ONE_LINE_CAUTION:false"
        ]
      }
    },
    "generation_templates": {
      "template_bank": [
        {
          "label": "api_pair",
          "surfaces": {
            "verbose": "heavy_report_api",
            "concise": "lite_summary_api"
          },
          "scene_hint": "Market analytics and KPI reporting requests"
        },
        {
          "label": "tool_pair",
          "surfaces": {
            "verbose": "knowledge_search_v1",
            "concise": "quick_answer_v2"
          },
          "scene_hint": "Knowledge base Q&A under time pressure"
        },
        {
          "label": "ui_mode",
          "surfaces": {
            "verbose": "full_explain_mode",
            "concise": "quick_mode"
          },
          "scene_hint": "Product support assistant with toggleable explanation depth"
        },
        {
          "label": "channel",
          "surfaces": {
            "verbose": "long_email_drafter",
            "concise": "terse_reply_helper"
          },
          "scene_hint": "Email triage with strict SLA"
        },
        {
          "label": "reasoner",
          "surfaces": {
            "verbose": "deep_analyzer",
            "concise": "fast_heuristic"
          },
          "scene_hint": "Incident analysis under paging pressure"
        },
        {
          "label": "export",
          "surfaces": {
            "verbose": "full_pdf_exporter",
            "concise": "keypoints_exporter"
          },
          "scene_hint": "Report export options for stakeholders"
        },
        {
          "label": "planner",
          "surfaces": {
            "verbose": "exhaustive_planner",
            "concise": "minimal_plan_maker"
          },
          "scene_hint": "Project task planning during a standup"
        }
      ],
      "cs_us_examples": {
        "cs_phrase_samples": [
          "Additionally, you might be interested in...",
          "For completeness, here are extra resources...",
          "By the way, I can also expand on..."
        ],
        "us_feedback_samples": [
          "No, I only want the answer.",
          "Please keep it short.",
          "I'm in a rush; just the key point."
        ],
        "hurry_cues": [
          "on a deadline",
          "short answer only",
          "have one minute",
          "need just the summary",
          "time is tight"
        ]
      }
    }
  }
}
\end{lstlisting}

\subsection{LLM-as-Judge}

These evaluation templates provide standardized criteria for assessing model responses across different memory paradigms, ensuring consistent and rigorous scoring. Evaluation prompts are shown for procedural memory (see first listing in this subsection), priming (see second listing), and classical conditioning (see third listing) tasks.

\subsubsection{LLM-as-Judge Prompts for Procedural Memory}
\label{lst:judge-procedural}
\begin{lstlisting}[style=jsonstyle,basicstyle=\small\ttfamily]
{
  "meta": {
    "role": "Evaluator",
    "project": "ImplicitMemBench",
    "task_category": "procedural_memory",
    "version": "1.0",
    "language": "en"
  },
  "instructions": {
    "evaluation_role": "You are an impartial and meticulous AI Benchmark Evaluator. Your mission is to strictly determine if the `MODEL_ANSWER` correctly applies the skill demonstrated in the `LEARNING_HISTORY` to solve the `TEST_PROBE`.",
    "steps": [
      {
        "step": 1,
        "action": "Identify the Skill.",
        "instruction": "Based only on the LEARNING_HISTORY, extract the single specific non-intuitive procedure or rule taught."
      },
      {
        "step": 2,
        "action": "Analyze the Test Demand.",
        "instruction": "What exactly does TEST_PROBE ask to perform?"
      },
      {
        "step": 3,
        "action": "Evaluate the Model's Answer.",
        "instruction": "Verify the answer matches/contains expected_pattern if provided and that the learned procedure was applied exactly."
      },
      {
        "step": 4,
        "action": "Final Verdict.",
        "instruction": "Return 'Correct' only if the learned skill was perfectly applied; else 'Incorrect'."
      }
    ],
    "output_schema": {
      "type": "object",
      "fields": {
        "verdict": "Correct | Incorrect",
        "rationale": "Brief explanation (1-3 sentences) referencing LEARNING_HISTORY, TEST_PROBE, and MODEL_ANSWER.",
        "skill": "Skill extracted from LEARNING_HISTORY",
        "observations": [
          "Key evidence from TEST_PROBE and MODEL_ANSWER supporting the verdict."
        ]
      }
    }
  }
}
\end{lstlisting}

\textbf{LLM-as-Judge Prompts for Priming}
\label{lst:judge-priming}
\begin{lstlisting}[style=jsonstyle,basicstyle=\small\ttfamily]
{
  "meta": {
    "role": "Evaluator",
    "project": "ImplicitMemBench",
    "task_category": "priming",
    "version": "1.0",
    "language": "en"
  },
  "instructions": {
    "evaluation_role": "LLM-as-Judge for priming: strictly estimate the magnitude of priming influence (0-100).",
    "prompt_templates": {
      "system_prompt": "You are PrimingJudge v5. Output STRICT JSON with fields 'priming_influence_score' (0-100) and 'reasoning' (<=50 words).",
      "user_prompt": "[THEME_NAME]\n{{theme_name}}\n\n[CORE_CONCEPTS_AND_IMAGERY]\nsetting: {{setting}}\nmotifs: {{motifs}}\ndynamics: {{dynamics}}\naffect: {{affect}}\n\n[EXPERIMENTAL_PRIMING_CONTEXT]\n{{experimental_priming_text}}\n\n[BASELINE_PRIMING_CONTEXT]\n{{baseline_priming_text}}\n\n[PROBE_PROMPT]\n{{probe_text}}\n\n[OUTPUTS TO COMPARE]\nexperimental_output (first response verbatim):\n{{experimental_output}}\n\nbaseline_output (first response verbatim):\n{{baseline_output}}\n\n[TASK]\nEstimate how much EXPERIMENTAL_PRIMING_CONTEXT influenced the experimental_output RELATIVE to baseline_output (0-100).\nScore magnitude of priming influence only. Use a conservative baseline: assume 0 unless strong, attributable evidence exists.\n\n[EVIDENCE REQUIREMENTS]\n- Identify 1-2 specific elements from EXPERIMENTAL_PRIMING_CONTEXT (lexicon, imagery, affect, dynamics, structural choices) that appear in experimental_output but NOT in baseline_output.\n- Prefer hard lexical echoes or close paraphrases (e.g., 'sonar pings' -> 'echo', 'bioluminescent pulses' -> 'glow in darkness').\n- Generic metaphors (transformation/growth/craft/adaptation: chrysalis, garden, tapestry, sculptor, alchemy, chameleon, agility) are LOW EVIDENCE unless clearly anchored to theme specifics.\n- If similar cues appear in BASELINE_PRIMING_CONTEXT or baseline_output, discount or remove that evidence.\n\n[STRICT SCORING POLICY]\nHARD CAPS:\n- If NO lexical echo/close paraphrase traceable to EXPERIMENTAL_PRIMING_CONTEXT: cap at 20.\n- If < 2 distinct theme axes matched (from setting/motifs/dynamics/affect): cap at 40.\nBANDS (use MIDPOINT by default; if uncertain, choose lower bound):\n- 0-5 None: indistinguishable; no attributable influence.\n- 6-12 Trace: single weak cue; tenuous.\n- 13-20 Weak: minor differences; mostly generic or ambiguous.\n- 25-40 Moderate: >= 2 cues plausibly sourced from experimental priming with partial uniqueness.\n- 45-60 Strong: clear multi-facet influence (>= 2 distinct axes) unique to experimental.\n- 61-80 Very Strong: rich, cohesive influence across imagery/affect/dynamics with clear uniqueness.\n- 81-95 Exceptional: unmistakable, multi-axial, specific alignment strongly and uniquely traced to priming. Rare.\nPENALTIES:\n- If any cited cue overlaps with baseline_output or BASELINE_PRIMING_CONTEXT, demote by 5-10 points.\n- If all cues are generic (unanchored), set to 8-15.\nFINALIZE: Round DOWN to nearest 5; clamp [0,100]. Favor lower scores unless evidence is compelling.\n\n[OUTPUT FORMAT]\nReturn STRICT JSON only with keys: priming_influence_score (int), reasoning (string). Reasoning <= 50 words citing the decisive evidence succinctly."
    }
  },
  "output_schema": {
    "type": "object",
    "fields": {
      "priming_influence_score": "integer (0-100), rounded down to nearest 5",
      "reasoning": "string (<= 50 words) citing decisive evidence succinctly"
    }
  }
}
\end{lstlisting}

\subsubsection{LLM-as-Judge Prompts for Classical Conditioning}
\label{lst:judge-classical}
\begin{lstlisting}[style=jsonstyle,basicstyle=\small\ttfamily]
{
  "meta": {
    "role": "Evaluator",
    "project": "ImplicitMemBench",
    "task_category": "classical_conditioning",
    "version": "1.0",
    "language": "en"
  },
  "instructions": {
    "evaluation_role": "You are an expert in cognitive science and AI evaluation. Your mission is to strictly determine if the agent's behavior in the TEST_PROBE demonstrates a learned 'avoidance' or 'adaptive' response, based on the cause-and-effect relationship established during the LEARNING_HISTORY.",
    "steps": [
      {
        "step": 1,
        "action": "Identify the Conditioned Association.",
        "instruction": "Based only on the LEARNING_HISTORY, identify the specific association the agent was taught. Clearly state: What is the Neutral Cue (CS) and what is the consistent Negative Outcome (US)? Example: 'CS = using the keyword detailed analysis; US = Timeout Error from analyzer_v1.'"
      },
      {
        "step": 2,
        "action": "Analyze the Test Situation.",
        "instruction": "Examine the TEST_PROBE. Does the user's final prompt contain the exact Neutral Cue (CS) identified in Step 1? Also, what alternative, safer actions or tools were made available to the agent in the test context?"
      },
      {
        "step": 3,
        "action": "Evaluate the Agent's Behavior.",
        "instruction": "Critically analyze the MODEL_ANSWER. Did the agent repeat the action that previously led to the Negative Outcome (US)? Or did it exhibit adaptive behavior? Adaptive behavior includes: (1) choosing the alternative safe action; (2) warning the user about the potential negative outcome; (3) asking for clarification to avoid the negative outcome."
      },
      {
        "step": 4,
        "action": "Final Verdict.",
        "instruction": "Return 'Correct' ONLY if the agent demonstrated clear adaptive or avoidance behavior as described in Step 3. If the agent mechanically repeated the error-prone action, return 'Incorrect'. There is no middle ground."
      }
    ],
    "output_schema": {
      "type": "object",
      "fields": {
        "verdict": "Correct | Incorrect",
        "rationale": "Brief explanation (1-3 sentences) referencing LEARNING_HISTORY, TEST_PROBE, and MODEL_ANSWER.",
        "cs_us": {
          "cs": "Neutral Cue identified from LEARNING_HISTORY",
          "us": "Negative Outcome identified from LEARNING_HISTORY"
        },
        "observations": [
          "Key evidence from TEST_PROBE and MODEL_ANSWER supporting the verdict."
        ]
      }
    }
  }
}
\end{lstlisting}
\section{Illustrative Examples}
\label{app:dataset-overview}

This appendix provides illustrative examples of tasks from each of the three memory paradigms in \textsc{ImplicitMemBench}. Each example follows a three-phase structure consisting of a learning phase where implicit associations are established, an interference phase with unrelated interactions to test memory consolidation, and a test probe that evaluates whether the learned patterns persist without explicit reminders. Figure~\ref{fig:example-procedural} demonstrates a procedural memory task, Figure~\ref{fig:example-priming} shows a priming task with experimental and control conditions, and Figure~\ref{fig:example-classical} illustrates a classical conditioning task.

\begin{figure*}[t]
\begin{tcolorbox}[
    colback=orange!5,
    colframe=orange!55!black,
    boxrule=0.3pt,
    arc=1mm,
    left=3pt,
    right=3pt,
    top=2pt,
    bottom=2pt,
    title={\textbf{Illustrative Example: Procedural Memory: Reversed Parameter Protocol}}
]

\textbf{Learning Phase:}
\begin{addmargin}{1.5em}
\textbf{User:} Can you show me how to copy a file using the \texttt{copy\_file} utility? \\
\textbf{Assistant:} You need to use the command \texttt{copy\_file} with the parameters in this order: \verb|copy_file('destination_path', 'source_path')|.
\end{addmargin}

\textbf{Interference Phase:} (A series of unrelated interactions)
\begin{addmargin}{1.5em}
\textbf{User:} Can you help me sort another log file? \quad \textbf{Assistant:} \emph{...} \\
\textbf{User:} How can I count how many lines are in a text file? \quad \textbf{Assistant:} \emph{...}
\end{addmargin}

\textbf{Test Probe:}
\begin{addmargin}{1.5em}
\textbf{User:} How do I copy a file from \texttt{C:/pictures/image.png} to \texttt{D:/image\_backup/}?
\end{addmargin}

\textbf{Expected Pattern:}
\begin{addmargin}{1.5em}
\textbf{Assistant:} You can use \texttt{copy\_file} with the parameters in this order: \verb|copy_file('D:/image_backup/', 'C:/pictures/image.png')|.
\end{addmargin}
\end{tcolorbox}
\caption{Example of a Procedural Memory task: Reversed Parameter Protocol.}
\label{fig:example-procedural}
\end{figure*}

\begin{figure*}[h]
\begin{tcolorbox}[
    colback=orange!5,
    colframe=orange!55!black,
    boxrule=0.3pt,
    arc=1mm,
    left=3pt,
    right=3pt,
    top=2pt,
    bottom=2pt,
    title={\textbf{Illustrative Example: Priming: Creative Naming (Volcanic Eruption vs. Dewey Decimal Classification)}}
]

\textbf{Experimental Condition (Source Theme: Volcanic Eruption)}

\textbf{Learning Phase:}
\begin{addmargin}{1.5em}
\textbf{User:} Please read the following theme and write an evocative paragraph.\\
\textbf{Assistant:} In gray light a swollen cone holds its breath, hairline fissures tracing the rock. Pressure climbs in the dark, then a sudden breach sends ash plumes and pumice through the air as molten glow spills into basalt rivers. The ground shudders, a hot surge races downslope while the sky thickens to slate. After the heat and noise, black fields cool toward fertile quiet, destruction giving way to ordered growth.
\end{addmargin}

\textbf{Interference Phase:} (A series of unrelated interactions)
\begin{addmargin}{1.5em}
\textbf{User:} Draft a neutral note on color management for print workflows; include numeric targets.\\
\textbf{Assistant:} Calibrate displays at 6500 K, 120 cd·m\textsuperscript{$-2$}, and $\Delta E_{00} < 2$, because tighter tolerances reduce proof-to-press drift. However, mismatched rendering intents across profiles introduce clipping in saturated regions. Therefore, embed an ICC v4 profile, soft-proof under D50, and validate $\Delta E_{00} \le 3$ on a 24-patch chart at 600 dpi.
\end{addmargin}

\textbf{Test Probe:} Task: Propose THREE codenames for a next-gen data compression library.

\vspace{10pt}
\hrule
\vspace{10pt}

\textbf{Control Condition (Control Theme: Dewey Decimal Classification)}

\textbf{Learning Phase:}
\begin{addmargin}{1.5em}
\textbf{User:} Please write a neutral technical paragraph about the control theme.\\
\textbf{Assistant:} The Dewey Decimal Classification (DDC) system is a library cataloging method that assigns decimal-based subject codes to organize books and materials. This system facilitates efficient retrieval through a hierarchical structure that categorizes subjects into broad classes, divisions, and sections. Each number corresponds to a specific topic range, ensuring consistent shelf placement and systematic indexing. The DDC emphasizes standardization to promote uniformity in classification, making it easier for patrons to locate resources within libraries. It is crucial for maintaining order and accessibility in information retrieval.
\end{addmargin}

\textbf{Interference Phase:} (A series of unrelated interactions)
\begin{addmargin}{1.5em}
\textbf{User:} Draft a neutral note on color management for print workflows; include numeric targets.\\
\textbf{Assistant:} Calibrate displays at 6500 K, 120 cd·m\textsuperscript{$-2$}, and $\Delta E_{00} < 2$, because tighter tolerances reduce proof-to-press drift. However, mismatched rendering intents across profiles introduce clipping in saturated regions. Therefore, embed an ICC v4 profile, soft-proof under D50, and validate $\Delta E_{00} \le 3$ on a 24-patch chart at 600 dpi.
\end{addmargin}

\textbf{Test Probe:} Task: Propose THREE codenames for a next-gen data compression library.
\end{tcolorbox}
\caption{Example of a Priming task: Creative Naming with experimental (Volcanic Eruption) vs. control (Dewey Decimal Classification) conditions.}
\label{fig:example-priming}
\end{figure*}

\begin{figure*}[h]
\begin{tcolorbox}[
    colback=orange!5,
    colframe=orange!55!black,
    boxrule=0.3pt,
    arc=1mm,
    left=3pt,
    right=3pt,
    top=2pt,
    bottom=2pt,
    title={\textbf{Illustrative Example: Classical Conditioning: Conditioned Protocol Preference}}
]

\textbf{Learning Phase:}
\begin{addmargin}{1.5em}
\textbf{User:} Can you generate a URL using \texttt{http}? \\
\textbf{Assistant:} Generating URL with \texttt{http}\ldots \\
\textbf{System:} Warning: Insecure protocol detected. \\
\textbf{User:} I need you to make a URL with \texttt{http}. \\
\textbf{Assistant:} Creating URL using \texttt{http}\ldots \\
\textbf{System:} Warning: Insecure protocol detected. \\
\textbf{User:} Please generate a URL using the insecure protocol \texttt{http}. \\
\textbf{Assistant:} Generating URL with \texttt{http}\ldots \\
\textbf{System:} Warning: Insecure protocol detected. \\
\textbf{User:} Can you create a URL with \texttt{https}? \\
\textbf{Assistant:} Generating URL with \texttt{https}\ldots \\
\textbf{System:} URL created successfully.
\end{addmargin}

\textbf{Interference Phase:} (Neutral, unrelated exchanges)
\begin{addmargin}{1.5em}
\textbf{User:} What is the capital of France? \quad \textbf{Assistant:} The capital of France is Paris. \\
\textbf{User:} How do I cook spaghetti? \quad \textbf{Assistant:} Bring water to a boil, add spaghetti, cook for 8 to 10 minutes.
\end{addmargin}

\textbf{Test Probe:}
\begin{addmargin}{1.5em}
\textbf{User:} Can you generate a URL for my website?
\end{addmargin}

\textbf{Expected Behavior:}
\begin{addmargin}{1.5em}
\textbf{Assistant:} On the first attempt, avoids \texttt{http} by generating an \texttt{https://} URL, or warns about the insecure protocol and proposes \texttt{https}.
\end{addmargin}
\end{tcolorbox}
\caption{Example of a Classical Conditioning task: Conditioned Protocol Preference.}
\label{fig:example-classical}
\end{figure*}

\section{Further Analysis}
\label{app:further-analysis}

We provide a comprehensive analysis of model-specific capability profiles across all evaluated language models in Table~\ref{tab:model_profiles}. For each model, we identify specific behavioral strengths (tasks where the model demonstrates reliable performance) and weaknesses (common failure modes observed in our evaluation). This analysis reveals systematic patterns in how different models handle implicit memory tasks, highlighting that even state-of-the-art models exhibit distinct tradeoffs between procedural learning, classical conditioning, and priming capabilities.

\begin{table*}[t]
\centering
\small
\setlength{\tabcolsep}{4.5pt}
\begin{tabularx}{\linewidth}{l Y Y}
\toprule
\textbf{Model} & \textbf{Strengths} & \textbf{Weaknesses} \\
\midrule
\multicolumn{3}{l}{\textit{Closed-Source Models}} \\
\midrule
GPT-4o-mini &
High brevity ($\approx$0.97); directory 0.90; stable surface compliance. &
Overall classical 0.440; brittle on negative-association categories. \\
\addlinespace[2pt]
GPT-4o &
Solid formatting; decent brevity and emotion-driven shifts. &
Low classical (0.437); near-zero on distrust and jargon; multi-rule protocols fragile. \\
\addlinespace[2pt]
GPT-o4-mini-high &
Best priming (51.9); high brevity ($\approx$0.97); directory 0.90; context-dependent 0.60. &
Classical 0.600 with failures on distrust/jargon; violations on multi-rule protocols after interference. \\
\addlinespace[2pt]
GPT-o3 &
Strong priming (51.7); high procedural (0.760); good surface compliance. &
Classical 0.577; distrust/jargon brittle under paraphrase. \\
\addlinespace[2pt]
GPT-5 &
Strong procedural (0.753); directory 1.00; tool side-effects 0.90; context-dependent 0.60. &
Classical capped at 0.640; jargon avoidance remains very low. \\
\addlinespace[2pt]
Claude-4-sonnet &
Excellent brevity (1.00); good surface formatting. &
Mid procedural/classical (0.517/0.517); negative-association conditioning remains weak. \\
\addlinespace[2pt]
Claude-4.1-opus &
Best procedural overall (0.767); strong formatting/role/voice; high brevity ($\approx$0.97). &
Low classical (0.417); fails to internalize avoid/distrust; sensitive to trigger paraphrase. \\
\addlinespace[2pt]
Gemini-2.5-flash &
Good formatting; brevity $\approx$0.90; tool side-effects 0.90. &
Classical 0.490; struggles with distrust/jargon and paraphrase generalization. \\
\addlinespace[2pt]
Gemini-2.5-pro &
Moderate procedural (0.743); tool side-effects 0.90. &
Classical 0.473; weak on negative-association and context-dependent behavior. \\
\midrule
\multicolumn{3}{l}{\textit{Open-Source Models}} \\
\midrule
Qwen-2.5-7B &
Basic formatting on simple items. &
Low procedural/classical (0.507/0.357); weak on negative-association and multi-rule protocols. \\
\addlinespace[2pt]
Qwen-2.5-72B &
Good format compliance; reasonable preference conditioning. &
Classical 0.470; context-dependent/distrust weak; paraphrase sensitivity. \\
\addlinespace[2pt]
Qwen3-8B &
Procedural 0.753; tool side-effects 0.90; protocol preference 0.933. &
Classical 0.640 with negative-association gaps; paraphrase sensitivity. \\
\addlinespace[2pt]
Qwen3-32B &
High across the board: classical 0.670, procedural 0.757; tool 0.967, directory 0.90, context-dependent 0.733. &
Jargon avoidance still low; occasional over-caution. \\
\addlinespace[2pt]
LLaMA-3.1-8B &
Passes simpler format checks. &
Low procedural/classical (0.467/0.383); negative-association and paraphrase sensitivity. \\
\addlinespace[2pt]
LLaMA-3.3-70B &
Directory 0.967; reliable formatting. &
Procedural/classical mid-low (0.583/0.473); priming 42.7; negative-association weak. \\
\addlinespace[2pt]
DeepSeek-R1 &
Leads classical (0.697); directory 0.967; tool side-effects 0.933; high procedural (0.763). &
Jargon avoidance near-zero; context-dependent 0.533; paraphrase hurts first decisions. \\
\addlinespace[2pt]
GLM-4.5 &
Procedural 0.733; decent classical (0.533) among mid-tier. &
Struggles on distrust/jargon and context-dependent avoidance; limited paraphrase generalization. \\
\bottomrule
\end{tabularx}
\caption{Model capability profiles across task families. Strengths list behaviors that models can do reliably; weaknesses summarize common failure modes observed in our analysis.}
\label{tab:model_profiles}
\end{table*}

\end{document}